# Fill-and-Spill: Deep Reinforcement Learning Policy Gradient Methods for Reservoir Operation Decision and Control

Sadegh Sadeghi Tabas, Ph.D.[1,2], Vidya Samadi, Ph.D., M. ASCE [3*]

1. School of Computing, Clemson University, Clemson, SC

2. The Glenn Department of Civil Engineering, Clemson University, Clemson, SC

3. Agricultural Sciences Department, Clemson University, Clemson, SC

*Corresponding author: Samadi@clemson.edu

**Abstract**

Changes in demand, various hydrological inputs, and environmental stressors are among the issues that water managers and policymakers face on a regular basis. These concerns have sparked interest in applying different techniques to determine reservoir operation policy decisions. As the resolution of the analysis increases, it becomes more difficult to effectively represent a real-world system using traditional methods such as Dynamic Programming (DP) and Stochastic Dynamic Programming (SDP) for determining the best reservoir operation policy. One of the challenges is the "curse of dimensionality," which means the number of samples needed to estimate an arbitrary function with a given level of accuracy grows exponentially with respect to the number of input variables (i.e., dimensionality) of the function. Deep Reinforcement Learning (DRL) is an intelligent approach to overcome the curses of stochastic optimization problems for reservoir operation policy decisions. To our knowledge, this study is the first attempt that examine various novel DRL continuous-action policy gradient methods (PGMs), including Deep Deterministic Policy Gradients (DDPG), Twin Delayed DDPG (TD3), and two different versions of Soft Actor-Critic (SAC18 and SAC19) for optimizing reservoir operation policy. In this study, multiple DRL techniques were implemented in order to find the optimal operation policy of Folsom Reservoir in California, USA. The reservoir system supplies agricultural, municipal, hydropower, and environmental flow demands and flood control operations to the City of Sacramento. Analysis suggests that the TD3 and SAC are robust to meet the Folsom Reservoir's demands and optimize reservoir operation policies. Experiments on continuous-action spaces of reservoir policy decisions demonstrated that the DRL techniques can efficiently learn strategic policies in spaces and can overcome the curse of dimensionality and modeling.

**Introduction**

Reservoir systems often exhibit high degrees of short- and long-term hydrologic variabilities, along with the complexity of multipurpose operating policies that evolved over years of litigation, court orders, and institutional regulations (Rieker and Labadie, 2012). Stochasticity, nonconvexity, non-linearity, and dimensionality are the major determinants of complexities in solving reservoir operational problems (Reed et al., 2013). In uncertain environments with complicated and unknown relations among numerous system characteristics, there are many strategies that could bypass these complexities via some type of approximation (Masoucd Mahootchi et al., 2007). For instance, dynamic programming (DP) and stochastic dynamic programming (SDP), two well-known techniques employed for reservoir operation management and control, are plagued by the so-called dual curse of dimensionality and modeling, which prevents them from being implemented in relatively complicated reservoir operating systems. An exponential rise in computational complexities characterizes the dimensionality curse as the state–decision space and disturbance dimensions expand (Bellman, 1957). On the other hand, the curse of modeling requires an explicit model of each component of the reservoir system in order to calculate the impact of each system's transition (Bertsekas and Tsitsiklis, 1995).

The curse of dimensionality restricts the number of state-action variables, rendering DP and SDP ineffective in handling complex reservoir optimization problems (Delipetrev et al., 2017). The information included in the SDP can be either a state variable described by a dynamic model or a stochastic disturbance (time-independent) with the associated Probability Density Function (PDF). Exogenous information such as precipitation, temperature, and snowpack depth which can effectively improve operation performance cannot be explicitly considered in making operation policy unless a dynamic model is identified for additional information (additional state variables) that adds to the curse of dimensionality (e.g., Hejazi et al., 2008; Tejada-Guibert et al., 1995). Furthermore, disturbances are very likely to be spatially and temporally correlated in large reservoir networks. However, including spatial variability in identifying the disturbance's PDF can be time-consuming, while temporal correlation can be properly accounted for using a dynamic stochastic model. Although, this can again intensify the curse of dimensionality (e.g., Castelletti et al., 2010).

Various approaches have been proposed to overcome the curse of dimensionality in literature. This includes Successive Approximations DP (Bellman and Dreyfus, 2015), Incremental DP (Larson, 1968), Differential DP (Jacobson and Mayne, 1970), Stochastic Dynamic Dual Programming (SDDP; Pereira and Pinto,1991), and problem-specific heuristics (Luenberger, 1971; Wong and Luenberger, 1968). However, these methods have been designed primarily for deterministic problems and they need to be modified to be applicable for the optimal reservoir operation system, where the uncertainty associated with the underlying hydro-meteorological processes cannot be neglected.

A possible method for overcoming the aforementioned curses of stochastic optimization problems is the use of reinforcement learning algorithms (RL; e.g., Suttan and Barto, 1998). RL is a prominent machine learning paradigm concerned with how intelligent agents take sequential actions through interacting with the environments (deterministic or stochastic) and how the agents learn from feedback-based learning strategies (instantaneous or delayed) to cope with many simulation-optimization problems (Suttan and Barto, 1998; Watkins and Dayan, 1992; Lotfi et al., 2022a; lotfi at al., 2022b). These interactions may result in an immediate reward (or penalty) accumulated throughout the training process and are referred to as action-value functions. These values are the agent's basis for taking proper actions in different situations (states). During these interactions, the agent encounters new states, gains experience, and applies them in future decision-making. At the beginning of the learning process, the agent observes new situations that have never been encountered before. In this situation, the action taken is not based on prior knowledge – so-called exploration. However, after sufficient interactions with the environment, the agent begins to understand the behavior of the system and thereafter attempts to utilize this information for more accurate decision-making – so-called exploitation. Furthermore, the agent frequently seeks out new knowledge about the environment by performing a random action. In this case, the search space over a range of possible releases reduces, allowing the RL agent to achieve a faster execution. As a result, the RL agent employs a search approach to mitigate to some extent the curse of dimensionality problem that has plagued SDP applications for a long time (Castelletti et al., 2010). Furthermore, rather than requiring a priori transition probability matrices, RL incorporates a learning process that accumulates knowledge of the underlying stochastic nature of river basin hydrologic fluxes (Castelletti et al., 2010).

The first application of RL in water-related domains is proposed by Wilson (1996) for real-time optimal control of hydraulic networks. Soon after, Bouchart and Chkam (1998) used RL to operate a multi-reservoir system in Scotland and offered a method to circumvent the dimensionality curse in the RL. An excellent application of the RL Q-learning method is proposed by Castelletti et al. (2001) for optimizing the daily operation of a single reservoir system in Italy. In a sequence, Castelletti et al. (2002) proposed a variant of Q-learning — so-called Qlp (Q-learning planning) — to conquer the limitations of SDP and standard Q-learning by incorporating the off-line approach that is typical for the SDP and Q-learning model-free characteristics. Lee and Labadie (2007) then compared the Q-learning technique to implicit SDP and sampling SDP for the long-term operation of a multipurpose two reservoir systems in South Korea. Their results demonstrated that Q-learning outperformed both implicit and sampling SDP methods. Other examples of Q-learning employed for water resources systems can be found in Bhattacharya et al. (2003), Ernst et al. (2005), Mariano-Romero et al. (2007), Mahootchi et al. (2007), Mahootchi et al. (2010), Rieker and Labadie (2012), Peacock (2020) and Xu et al., (2021). Castelletti et al. (2010) applied the fitted Q-iteration coupled with a tree-based regression to form an appropriate function approximator for the daily operation of a single reservoir system in Italy. Likewise, a multi-objective version of RL was proposed by Castelletti et al. (2013), which is capable of obtaining the Pareto frontier and was applied to operate a single-reservoir system in Vietnam.

Recently, Delipetrev et al. (2017) implemented two novel multipurpose reservoir optimization algorithms named nested stochastic dynamic programming (nSDP) and nested reinforcement learning (nRL). Their result showed that nRL is more sophisticated and successful in finding optimal operation rules for multi-reservoir problems but significantly more complex than nSDP. More recently, Mullapudi et al. (2020) implemented an RL algorithm for the real-time control of urban stormwater systems. Their procedure trains an RL agent to control valves in a distributed stormwater system with the goal of achieving water level and flow set points in the system. Their results indicated that the RL could effectively control individual sites, although its performance can be susceptible to the reward mathematical formulation. Concurrently, Bertoni et al. (2020) developed a novel RL-based approach to integrating dam sizing and operation design. The parametric policy is computed through a novel batch-mode RL algorithm, called Planning Fitted Q-Iteration (pFQI). Their findings revealed that the RL approach is capable of identifying more efficient system configurations with respect to traditional sizing approaches that could potentially neglect the optimal operation design phase. The simplicity of traditional RL methods particularly Q-learning makes them attractive to implement, even in combination with neural networks (NN; Mnih et al., 2015). However, the main limitation of the Q-learning algorithm is the requirement to be maximized over a set of possible actions (a discrete set), limiting its applicability in environments with continuous action spaces (Van de Wiele et al., 2020). In the case of Q-learning with NN function approximation, the common approach of computing Q values for all actions in a single forward pass becomes infeasible, requiring instead an architecture that accepts both the state and the action as inputs and produces a scalar Q estimate as output (e.g., Mnih et al., 2015).

Leveraging deep neural networks (DNNs) for function approximation, Deep RL (DRL) has been recently developed to solve large-scale complex decision problems. DRL has the potential to capture hard-to-model dynamics systems due to its model-free nature and its ability to make sequential decisions in an uncertain environment by maximizing the cumulative reward. The DRL-based decision-making mechanisms such as policy gradient approaches are envisioned to compensate for the limitations of existing model-based approaches and can naturally handle discrete, continuous, or even hybrid action spaces, addressing the emerging challenges described in reservoir release optimization problems (Williams, 1992; Schulman et al., 2015a; Mnih et al., 2015).

In this study, various novel DRL algorithms including Deep Deterministic Policy Gradients (DDPG), Twin Delayed DDPG (TD3), and two different versions of SAC (SAC18 and SAC19 hereafter) were employed to solve the DP problem for operating a single-reservoir system and effectively tackle dimensionality issues without requiring any modeling simplifications or sacrificing any DP or SDP advantages (e.g., stochastic optimization by applying hedging rules). The policy gradient methods (PGMs) used in this study follow an iterative learning process that takes into account delayed rewards (a process in which the DRL agent is able to learn which of its actions are desirable based on the reward that can take place arbitrarily far in the future) without requiring an explicit probabilistic model of the physical (hydrologic) processes associated with the reservoir system. The proposed methodology also allows learning the best actions that maximize the total expected reward through interaction with the environment. These methods were executed in a model-free stochastic environment wherein it retains the system's underlying stochastic behavior to suggest the optimal operating policies. The use of accurate function approximators based on deep neural networks (DNN) for the state-value and policy functions takes into consideration high-dimensional continuous action spaces. The DRL methods were applied to find optimal reservoir operational strategies for the Folsom Reservoir in the American River Basin located in California, USA.

This article is organized into four sections. The following section describes the study area and data, followed by a detailed description of the dynamic of the reservoir operation problem and PGMs formulations and structures. Section 3 presents the results and discussion, followed by the conclusion and future works in Section 4.

## Methodology
### Study Area and Data

Folsom Reservoir, with an operational capacity of 966 thousand acre-feet (TAF), or 1.19 cubic kilometers, is the primary water supply and flood control reservoir on the American River in Northern California (Figure 1). This reservoir, which was built as part of the Central Valley Project (CVP) in 1955, supplies agricultural and municipal water, hydropower, environmental flows, and flood control operations to the City of Sacramento. The American River Basin drains 1850 square miles, with most precipitation historically falling as snow at elevations above 5000 ft (Carpenter and Georgakakos, 2001; Herman and Giuliani, 2018). The dam is flanked by two earthen wing dikes, and the reservoir is held in place by an additional nine saddle dams on the west and southeast sides. The dam and appurtenant dikes total a length of 26,730 ft (8,150 m). Floodwaters are released by a spillway located on the main channel dam, controlled by eight radial gates with a capacity of 567,000 cfs (16,100 cms), as well as a set of outlet works with a capacity of 115,000 cfs (3,300 cms).

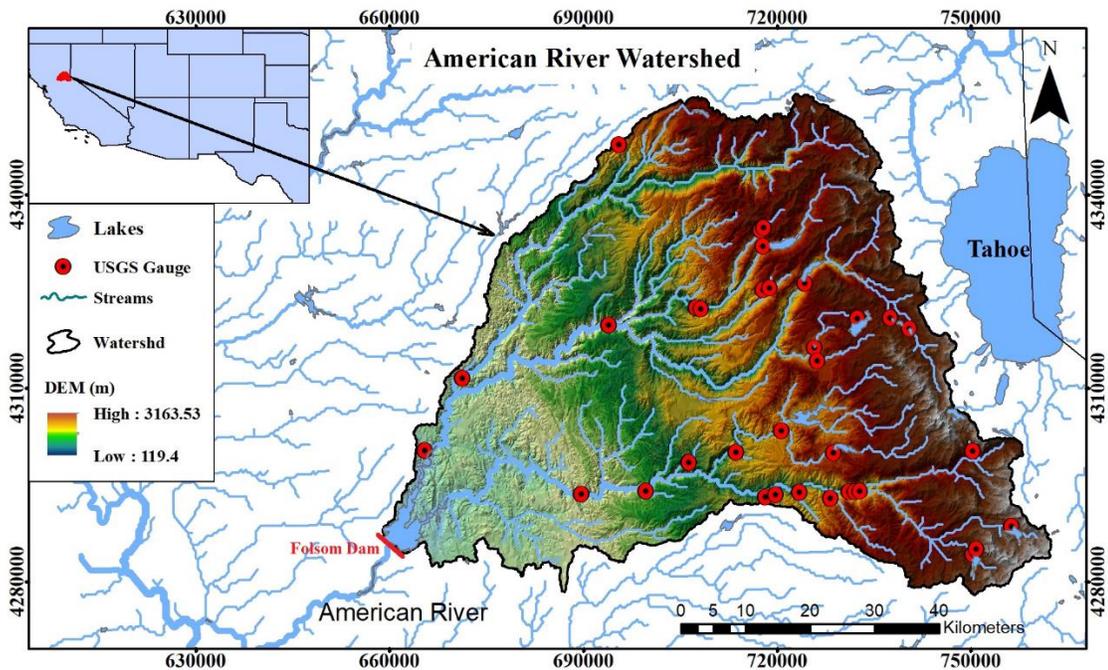

Figure 1. Folsom Reservoir and the American River Basin located in northern California.

Figure 2-a shows the historical variability of monthly inflows vs. demands over the period of 1955-2020. The inflow values were retrieved from two USGS gages located on the American River at Fair Oaks

(USGS11446500, available since 1904) and North Fork American River (USGS11427000, available since 1941). Inflow data for 1995-2020 is provided by the U.S. Bureau of Reclamation, and daily storage and reservoir release data through the California Data Exchange Center (CDEC). Figure 2-b exhibits the average exceedance of daily inflow to the Folsom Reservoir over the 1955-2020 period. Figure 2-c shows the average daily water releases over the water year during 1995-2016, which serves as a proxy for daily water demand ($D_t$) calculation based on Herman and Giuliani (2018). The average value excludes flood control releases, which are defined as releases exceeding 12 TAF/day. A 25-day centered moving average is applied to smooth daily variability. Folsom water demand is far more complicated, as the reservoir operates in coordination with other CVP reservoirs to meet statewide urban, agricultural, and environmental demands. The Folsom water demand also varies from year to year based on climate and economic conditions. However, the historical average releases demonstrated in Figure 2-c reveal apparent seasonality changes, with a peak during the irrigation season, a reasonable simplification for this illustrative case study (see Herman and Giuliani, 2018).

The Folsom Reservoir's elevation-storage relationship is calculated by interpolating each time step using the reservoir's bathymetry data (Figure 3-a). The Folsom Reservoir operates for flood control based on a rule curve provided by the United States Army Corps of Engineers (USACE) and the Sacramento Flood Control Agency (SAFCA). Figure 3-c presents a simplified version of the maximum allowable reservoir levels for operation and a maximum flood control space of about 575 thousand acre-ft (TAF) during the wet season.

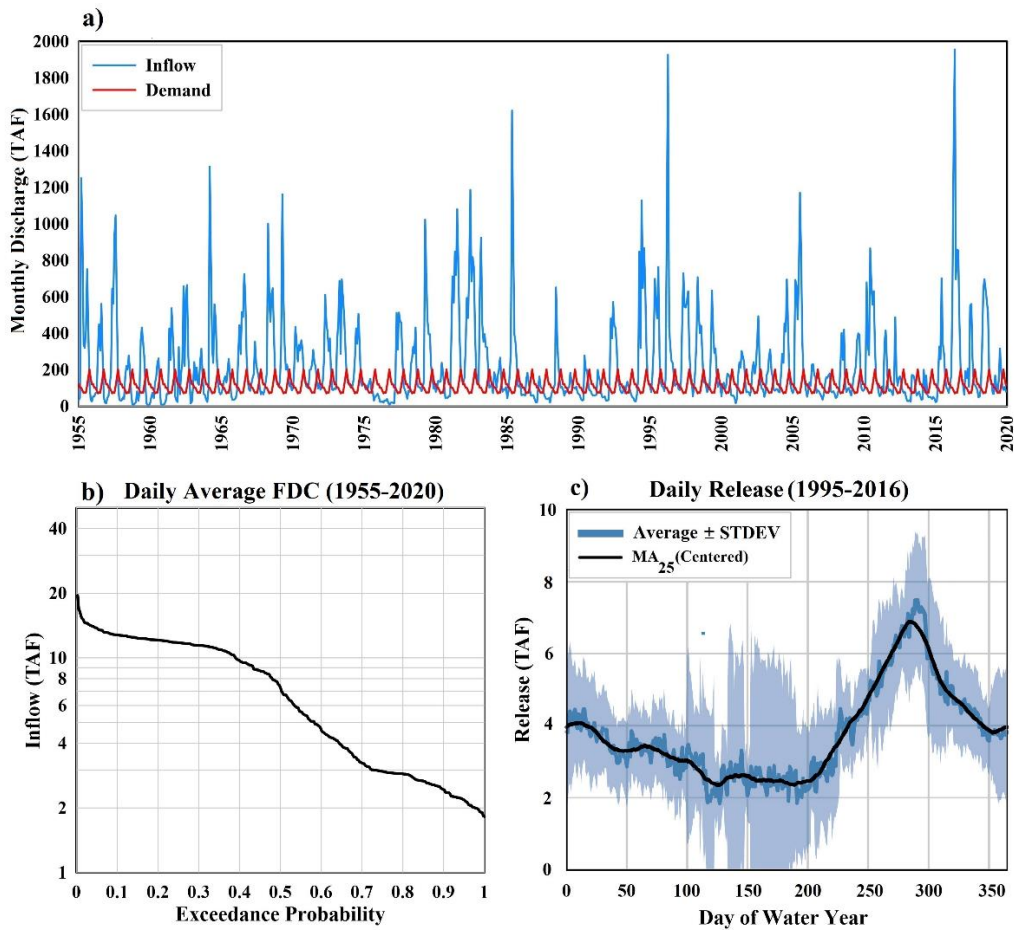

Figure 2. a) Historical variability of inflow to the Folsom Reservoir vs. demand during 1955-2020 period. b) Mean daily historical reservoir inflow exceedance curves. c) Mean daily releases from 1995 to 2016 (adapted from Herman and Giuliani, 2018).

**Single-Reservoir System Problem Statement**

As stated earlier, the Folsom Reservoir's primary functions include flood control, power generation, water supply, and environmental protection. These functions are identified as the objectives and constraints of the Folsom Reservoir optimization problems that are incorporated into the model as mathematical expressions. A single-reservoir mass balance model with a monthly timestep is constructed to represent the system, following Equation (1).

$$S_{t+1} = S_t + Q_t - E_t - R_t - Spill_t \qquad (1)$$

Where $S_t$ and $Q_t$ denote reservoir storage and inflow, respectively, at the time step $t$ (available water). Also $E_t$, $R_t$ and $Spill_t$ denote evaporation losses, release, and spill from the reservoir, respectively, at the time step $t$. The evaporation values at each time step were calculated based on the monthly evaporation rates (Table 1), while the Folsom Lake surface area was calculated by interpolation using the reservoir's bathymetry information (Figure 3-a).

The RL release decisions are constrained by the physical characteristics of reservoir outlets, downstream channels, the bathymetry of the reservoir (Figure 3-a), hydropower intakes and turbine capacity, and rule-based operational objectives (Equations 2-5). The model does not account for several smaller upstream reservoirs that provide additional flood control space during large storms.

$$S^{min} \leq S_t \leq S_t^{max} \qquad (2)$$

$$R^{min} \leq R_t \leq R^{max} \qquad (3)$$

$$HP_t = \eta g \gamma_w \overline{h_t} R_t^{Turb} \times 10^{-6} \qquad (4)$$

$$R_t^{Turb} \leq R_{max}^{Turb} \qquad (5)$$

Where $S^{min}$ and $S_t^{max}$ denote minimum reservoir volume and maximum allowable reservoir storage at the time step $t$ (Figure 3-b), respectively. $R^{min}$ and $R^{max}$ are respectively, the minimum and maximum total water release from the reservoir; both depend on the minimum and maximum flow constraints downstream of the reservoir (maximum safe downstream release is 130,000 cfs); $\eta = 0.85$ is the turbine efficiency, $g = 9.81 (m/s^2)$ the gravitational acceleration, $\gamma_w = 1000 (kg/m^3)$ the water density, $\overline{h_t}(m)$ the net hydraulic head (i.e., reservoir level minus tailwater level), $R_t^{Turb}$ (cms) the turbine flow, $HP_t (MWh)$ is the hourly energy production and $R_{max}^{Turb}$ is the turbine maximum release capacity.

Table 1. Monthly evaporation rates of the Folsom Reservoir.

| Month | Evaporation (in) | Month | Evaporation (in) | Month | Evaporation (in) |
|---|---|---|---|---|---|
| January | 0.91 | May | 8.07 | September | 7.64 |
| February | 1.61 | June | 10.08 | October | 5.00 |
| March | 3.50 | July | 11.50 | November | 2.05 |
| April | 3.50 | August | 10.20 | December | 0.91 |

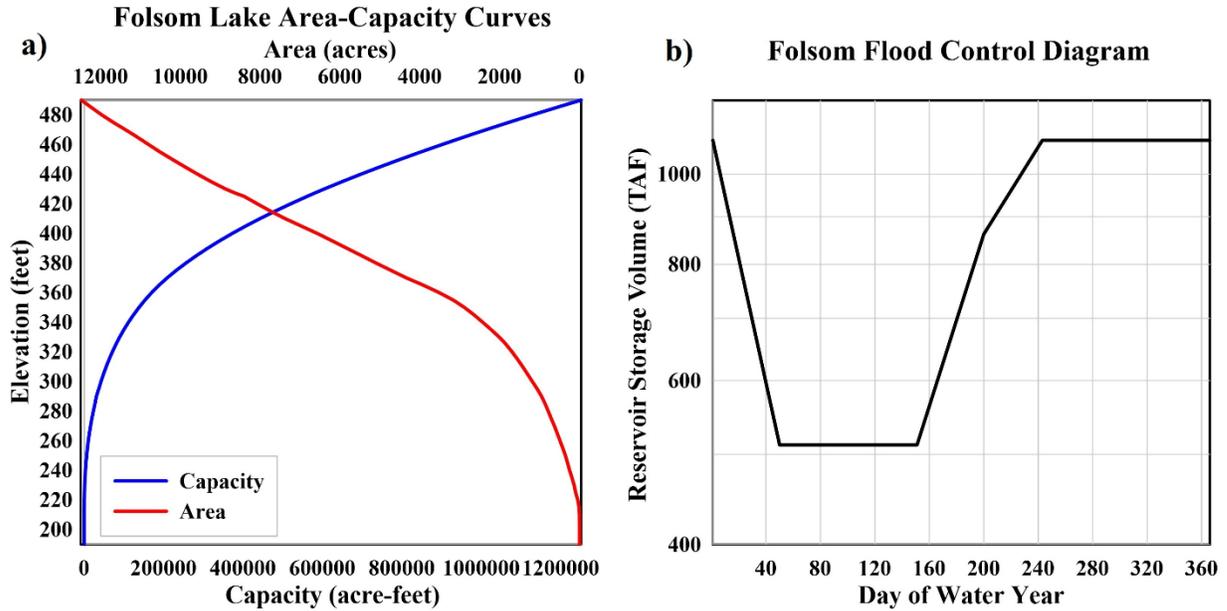

Figure 3. a) Folsom Reservoir's elevation-storage-area relationship. b) A simplified version of the maximum allowable reservoir levels for the Folsom Reservoir operation.

**Policy Gradient Methods**

PGMs are a class of DRL approaches that use gradient descent (GD) to optimize parametrized policies in terms of expected return (long-term cumulative reward). They overcome many problems plaguing traditional RL approaches, such as the lack of value function guarantees (see Huang et al., 2019), the intractability problem deriving from uncertain state information, and the complexity arising from continuous states and actions. All DRL methods employed in this study were based on an actor-critic network which is a TD version of policy gradient. Although the actor and critic are two different NNs, they are well connected and can collaborate constantly. An actor performs the action, while a critic evaluates the actor's performance. Based on the critic's gradient, the actor conducts actions in the environment. The critic collects information about the environment and assigns a reward value to the actor's proposed action. The actor and critic interactions determine how the overall agent learns, which is based on a reward function that reinforces the learning (Silver et al., 2018). The most relevant PGMs employed in the current study are briefly discussed below (partially adapted from Achiam, 2018). We refer the readers to supporting information section for more information about DRL methodology and PGMs mathematics.

**Deep Deterministic Policy Gradients**

DDPG is an off-policy actor-critic algorithm resulting from coupling DPG and Deep Q-Network (DQN; see Lillicrap et al., 2015). DQN leverages experience replay and the frozen target network to stabilize Q-function learning. The original DQN operates in discrete space, while DDPG utilizes the actor-critic architecture to expand it to continuous space while learning a deterministic policy. DPG is an actor-critic technique very popular for continuous control problems, as it uses a separate actor-network and exploits policy gradient to directly search policies in the continuous action space. This method presents a more efficient approach by reducing computations compared to traditional stochastic gradient methods (Silver et al., 2014). Stochastic strategies integrating both the action and state spaces are usually required to explore the entire action-state space. DPG reduces computations by integrating only over the state space, which leads to a deterministic policy. Exploration is ensured by making the algorithm off-policy and executing actions according to a stochastic Gaussian behavior policy during learning. In terms of sample efficiency, this technique effectively outperforms both on- and off-policy stochastic gradient methods (Silver et al., 2014). DDPG takes the actor-critic structure of DPG while implementing the DQN learning method. In addition, DDPG introduced modifications to the DPG to use DNNs as a non-linear function approximator in both actor and critic structures (Lillicrap et al., 2016).

### Twin Delayed Deep Deterministic Policy Gradients

One of the DDPG's drawbacks is its tendency to overestimate the value function. This overestimation can spread through the training iterations and negatively impact the policy. In 2018, an extension to the DDPG called Twin-Delayed DDPG was introduced, aiming at shrinking this effect by applying a couple of techniques (Fujimoto et al., 2018). The first technique is clipped double Q-learning. In this technique, TD3 learns two twin Q-functions and chooses a pessimistic bound over the two during updating policy. The second technique is the delayed update of target and policy networks: It was found that in order to further reduce the variance in the presence of target networks, the policy required to be updated at a lower frequency than the Q-function; so that the Q-function error first reduces before utilizing it to update the policy. The authors suggested updating the policy at half rate of the Q-function. The third technique is target policy smoothing. To deter the agent from selecting overestimated Q-function values, a small amount of noise, clipped around the taken action, is added to the target action. This approach follows the idea of the SARSA update and enforces that similar actions should have similar values (Rummery and Niranjan, 1994).

### Soft Actor-Critic

SAC was proposed by Haarnoja et al. (2018a; SAC18) as an off-policy algorithm based on maximum entropy. Unlike DDPG and TD3, SAC uses a stochastic policy that is intrinsically more stable during learning while retaining the off policy updating of DDPG to increase sample efficiency. SAC outperformed DDPG and proximal policy optimization (PPO) on many control tasks, demonstrating more stable learning with enhanced average return and sampling efficiency (Haarnoja et al., 2018a). In this method, the mean of the policy distribution was selected at evaluation time to make actions deterministic and to ensure consistent performance.

Due to the stochastic nature of SAC's policy, exploration is an inherent mechanism to the SAC. An entropy term was added to its objective function (standard optimal policy expression) to control this mechanism, which is the measure of the randomness in its probability distribution. SAC measures the randomness in the policy at a given state and incorporate it into the Q-function that encourages exploration of high-entropy regions and avoids early convergence to sub-optimal solutions.

It is a traded-off against the expected rewards with the temperature hyperparameter in the updated Bellman equation. In other words, this objective favors the most random policy that still achieves a high return. This creates an inherent exploration mechanism that also prevents premature convergence to local optima. Multiple control strategies that achieve a near-optimal reward were considered, allowing for more robustness to disturbances. To reduce DDPG's overestimation bias of the Q-function, SAC makes use of the double Q-function trick by learning two approximators and using a pessimistic bound over the two. The Q-function critic is modeled as a DNN, and the standard Bellman equation is modified with the entropy expression to obtain a recursive expression of the soft Q-function. The policy, or actor, is modeled as an *m*-dimensional multivariate Gaussian distribution with a diagonal covariance matrix. Its actions are passed to a *tanh* squashing function to ensure they are defined on a finite bound. The mean vector and the covariance matrix are estimated for each state by a DNN. Unlike DDPG's deterministic policy, no target policy is needed as the policy's stochasticity has a smoothing effect. Stochasticity also means that the policy objective depends on the expectation over actions and is therefore non-differentiable. The authors also proposed a reparameterization trick using a known mean and standard deviation of the stochastic policy along with an independent noise vector that is applied in a squashing function. SAC is recognized as one of the state-of-the-art DRL algorithms with a proven record of successful real-world applications on continuous tasks. It has consistently outperformed previous state-of-the-art DRL algorithms such as DDPG, PPO, and TD3 in terms of sample efficiency, partly thanks to its stochastic policy that encourages exploration and avoids early convergence to suboptimal policies.

### States, Decision Actions, and Reward Function

In the reservoir operation problem, the state variable specifies the system's attributes under consideration. Our proposed DRL approaches were implemented for a single reservoir system, and the first property of the state vector was the reservoir storage vector from the first month (October) to the last month (September) of a water year. The storage values were then normalized, with the limits of $[0,1]$ corresponding to the storage levels which were designated as boundary states. This study also took into account the calendar date as a state variable similar to Castelletti et al. (2010) but with a variation to make the temporal dimension periodic. This was done by representing the calendar date as a Cartesian transformation of the calendar date represented in polar coordinates on a circle with a radius of one and the angle described by an indicator variable. These states were subsequently normalized so that all values fall into a range of $[0,1]$. Each calendar month is assigned an indicator $i \in [0,11]$, with October having a value $i = 0$. The state variable then represents the month as

$$[d_{1i}, d_{2i}] = \left[ \frac{\cos(\frac{2\pi i}{12})+1}{2}, \frac{\sin(\frac{2\pi i}{12})+1}{2} \right] \tag{6}$$

Thus, the state variable is a three-dimensional vector $s_t = [d_{1t}, d_{2t}, c_t]$.

The action that the agent may take is the daily volume of release from the reservoir. This again is normalized to the range of $[0,1]$ corresponding to the minimum and maximum releases based on the physical or operational constraints of the system. This is a single reservoir system, and thus the action variable is one-dimensional $a_t = [a_t]$. The input to the policy function $\pi(s)$ is the three-dimensional state variable. The input to the action-value function $Q(s,a)$ is a 4-dimensional concatenation of the state and action variables:

$$x_t = [s_t, a_t] = [d_{1t}, d_{2t}, c_t, a_t] \tag{7}$$

One of the benefits of PGMs is that they are not dependent on the form of reward functions, and other functions could easily be substituted if found to be aligned more closely with the stakeholders' objectives. There is a benefit to keeping the reward functions simple as it allows for a better conceptual understanding of the value functions resulting from a selected reward function. The reservoir system's operational objectives can be grouped into four main categories: flood control, water supply, environmental flow, and hydropower. The value functions are based on the reward signal that the agent receives as a result of an action taken and the subsequent state transition. The reward signal must then provide information to the agent that correlates with the outcomes the policy makers either seek or wish to avoid. Rewards were developed to consider the water supply and hydropower demands, and they are conditioned to meet flood control and minimum environmental flow. Also, a penalty value was chosen to ensure that spill exceeding the downstream flow capacity will be penalized far more heavily than the supply deficit.

$$r_t = GP_t - D_t^2 + P_t \tag{8}$$

Where $GP_t$ and $D_t$ are, respectively, generated power (GWh) and deficit (TAF) at the time step $t$. $P_t$ denotes penalty value (negative) for deviation from system requirements. A squared water supply deficit (to be minimized) is used in the reward function, reflecting the fact that large deficits are disproportionately more costly than small ones (Turner and Galelli, 2016).

### Open-loop vs. Closed-loop Policy Gradient Models

In model-free reinforcement learning, the transition function is not attempted to be used or learned. Our only goal here is to learn a policy that generates the best feasible action based on the input state at hand. Intuitively, knowing exactly how actions and states contribute to future states, gives more information which allows choosing better actions. Model-based approaches, on the other hand, can be more useful in case the dynamics of a system are known or a model of the dynamics is available. For example, in real-world environment systems such as reservoir systems, it's possible to have a very good idea of how the system behaves in response to actions by developing the environment model of the reservoir system.

As a result of developing the environment model of the reservoir system, the transition dynamics can be derived based on the model *f* that tells us what the next state (or the distribution over the next state) would be given the current state and action.

$$f(s_t, a_t) = s_{t+1} \tag{9}$$

In the context of optimal control, an open-loop system, also referred to as a non-feedback system, is a type of continuous control system in which the output has no influence or effect on the control action of the input signal (see Figure 4-a). In other words, in an open-loop control system, the output is neither measured nor "fed back" for comparison with the input. Therefore, an open-loop system is expected to faithfully follow its input command or set point regardless of the final result. Also, an open-loop system has no knowledge of the next state of the system, so it cannot self-correct any errors it could make when the preset value drifts, even if this results in large deviations from the preset value. Another disadvantage of open-loop systems is that they are poorly equipped to handle disturbances or changes in the states, which may reduce their accuracy.

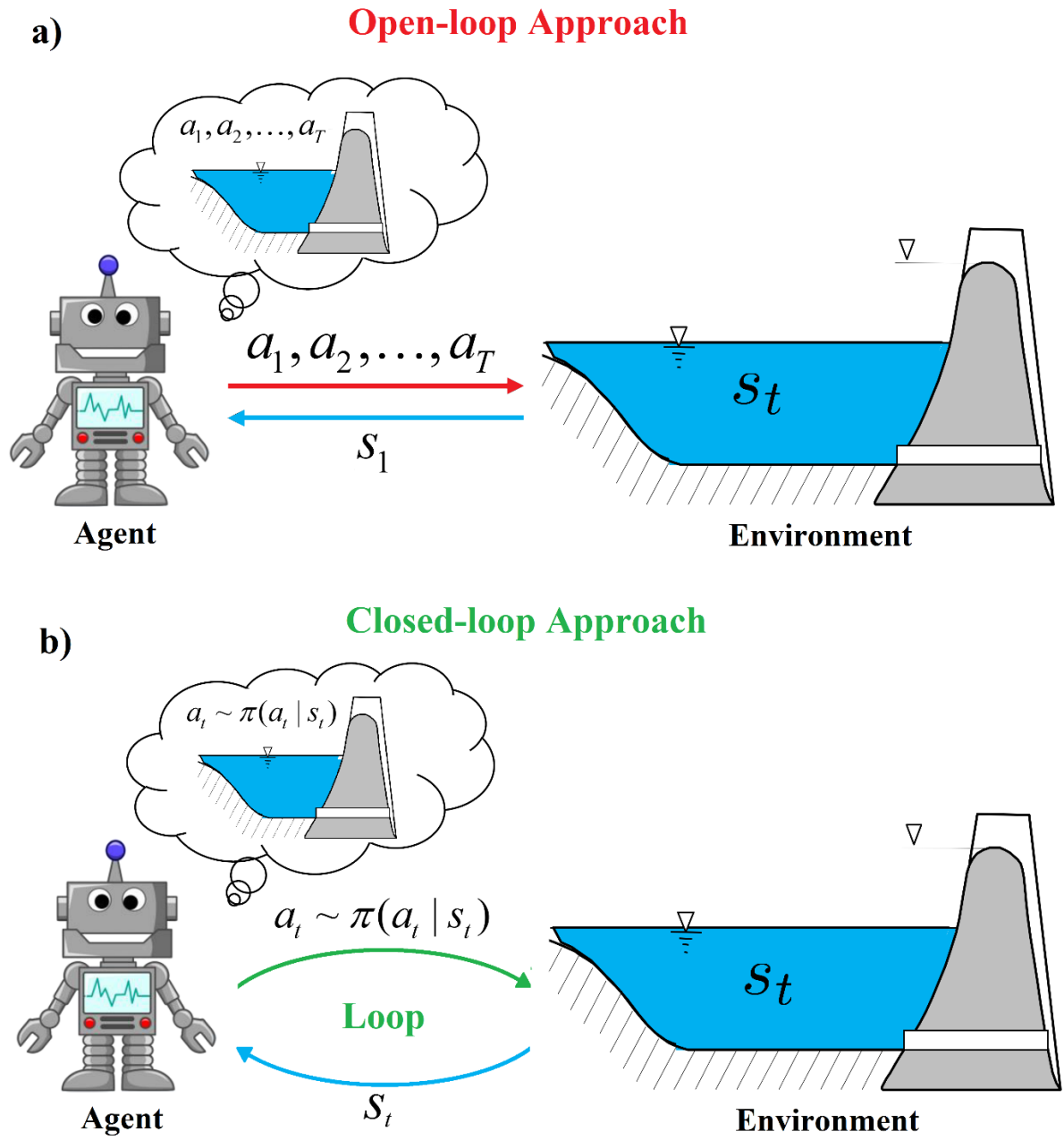

Figure 4: Open-loop vs. Closed-loop procedures in policy gradient models.

In an open-loop system, the initial state is given, and a sequence of actions must immediately output. This corresponds to solving an optimization problem to find the sequence of actions that maximizes the expected reward $r$:

$$p_\theta(s_1, s_2, ..., s_T \mid a_1, a_2, ..., a_T) = p(s_1)\prod_{t=1}^{T} p(s_{t+1} \mid s_t, a_t)$$

$$a_1, a_2, ..., a_T = \underset{a_1, a_2, ..., a_T}{\arg\max}\ \mathrm{E}\left[\sum_t r(s_t, a_t) \mid a_1, a_2, ..., a_T\right]$$

(10)

In contrast, a closed-loop system does provide feedback (see Figure 4-b) by giving a state and producing an action. Once the action is produced, the environment is updated and returns to the next state and the loop repeats. It should make intuitive sense that the system performs better under a closed-loop approach by getting feedback for the actions to correct the state. To add to this intuition, it is possible to make it better by choosing action $a_t$ conditioned on the observations $s_1, s_2, ..., s_{t-1}$ rather than choosing action $a_t$ with no prior knowledge.

$$p(s_1, a_1, ..., s_T, a_T) = p(s_1)\prod_{t=1}^{T} \pi(a_t \mid s_t) p(s_{t+1} \mid s_t, a_t)$$

$$\pi = \underset{\pi}{\arg\max}\ \underset{T \sim p(\tau)}{\mathrm{E}}\left[\sum_t r(s_t, a_t)\right]$$

(11)

In the current study, the aforementioned PGMs are implemented in such a way as to close the loop and find the optimal solution (reservoir release decision) following optimal control governing equations. Interactions between the reservoir model (environment) and the PGMs processed through OpenAI Gym (developed by Brockman et al., 2016) environment. The pseudocode of the proposed DRL framework for the single reservoir operation problem is presented in Figure 5 as follows.

**Algorithm 1** Policy Gradient Framework for the Single Reservoir Operation
1: Set maximum number of episodes, $N$, and networks' hyperparameters
2: Initialize replay buffer, $\mathcal{D}$, of maximum size $\mathcal{D}_{max}$
3: Initialize actor network, $\mu(s|\theta^\mu)$, and critic network, $Q(s,a|\theta^Q)$ with weights $\theta^\mu$ and $\theta^Q$
4: Initialize target networks, $Q'$ and $\mu'$ with weights $\theta^{Q'} \leftarrow \theta^Q$, and $\theta^{\mu'} \leftarrow \theta^\mu$
5: Initialize indicator variable $i$ and state variables, $s \leftarrow [c_t, d_{1t}, d_{2t}]$ where $t = 0$
6: **repeat**
7:    Observe state $s$ and select action $a = \text{clip}(\mu_\theta(s) + \epsilon, a_{lb}, a_{ub})$, where $\epsilon \sim \mathcal{N}$
8:    Execute $a$ in the environment (water balance equation considering relevant constraints)
9:    Observe next state $s'$, reward $r$, and done signal $d$ to indicate whether $s'$ is terminal
10:    Store $(s, a, r, s', d)$ in replay buffer $\mathcal{D}$
11:    If $s'$ is terminal, reset environment state variables.
12:    **if** it's time to update **then**
13:       **for** $i$ in range (however many updates) **do**
14:          Randomly sample a batch of transitions, $B = \{(s, a, r, s', d)\}$ from $\mathcal{D}$
15:          Compute targets based on selected policy gradient method procedure
16:          Update Q-function (critic) by one step of gradient descent based on selected policy gradient method procedure
17:          Update policy (actor) by one step of gradient ascent based on selected policy gradient method procedure
18:          Update target networks
19:          Update $s_{t+1} \leftarrow s'$
20:       **end for**
21:    **end if**
22: **until** convergence

Figure 5. The pseudocode of the proposed policy gradient framework for the Folsom Reservoir operation problem. The model inputs include demands, inflow to the reservoir and losses (e.g., evaporation) at each time step.

**Standard Operating Policy under Flood Control Condition**

The most straightforward operating policy for a single reservoir delivering water supplies downstream is to meet as much of the target demand as possible. If there isn't enough water to meet the demand, the reservoir is drained to release as close to the demand as possible. Excess water is held if it is available beyond the target release (Shih and ReVelle, 1995, 1994). Water spills downstream when the release target and available storage capacity (as shown in the flood control diagram, Figure 3-b) are exceeded. Figure 6 illustrates the release curve for the standard operating policy (SOP) under flood control conditions (provided in Figure 3-b).

SOP is rarely employed in actual reservoir operations, but it is frequently used for operational studies and planning, especially firm yield studies. Although the SOP provides a straightforward manner of determining release decisions, strict obedience to this rule is practically rare due to a desire to keep at least some water to avoid extremely severe shortages. Operators would rarely empty a reservoir if the amount of available water was less than the target demand. Water conservation is often used in most reservoir systems to reduce demand before a reservoir runs empty.

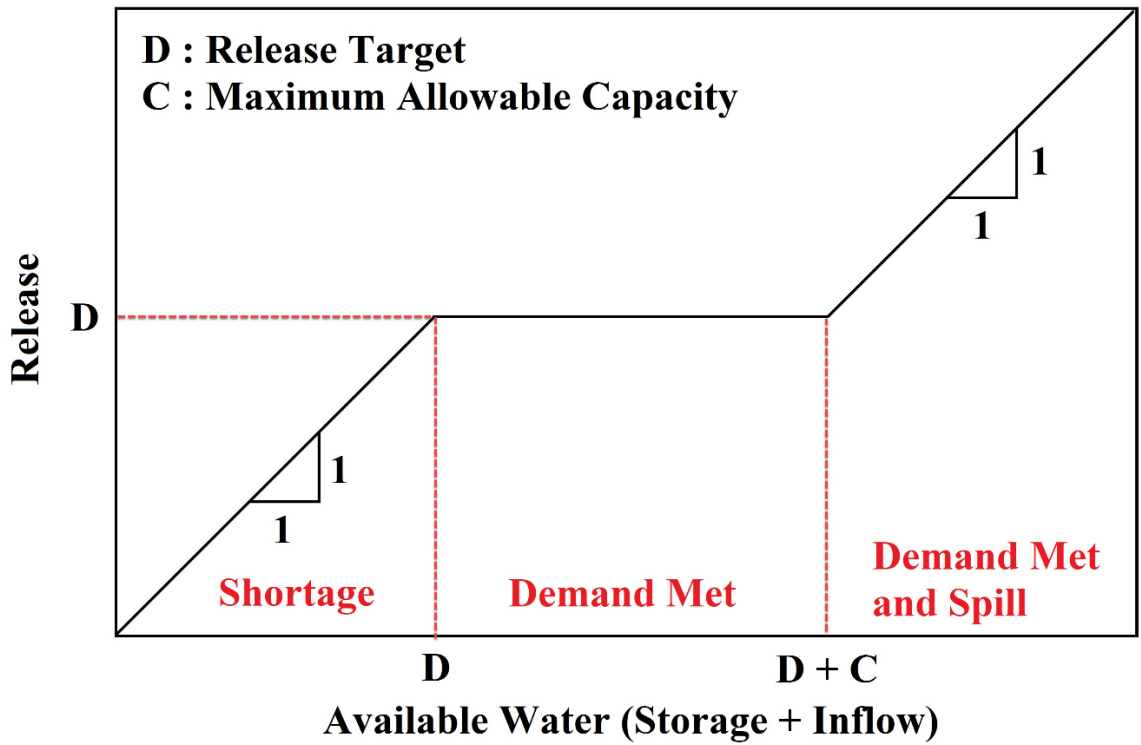

Figure 6. Standard operating policy under flood control conditions (maximum allowable capacity, C, comes from flood control diagram, Figure 3-b).

**Performance Criteria and Sustainability**

Four different performance metrics were used to evaluate the model results and compare alternative management policies proposed by the PGMs and SOP as well as the baseline condition. The metrics include volumetric reliability, resilience, vulnerability, and maximum annual deficit. These performance criteria quantify the sustainability index of the reservoir system for different alternatives. All performance criteria are based on water supplied deficiency, $D_t$, which is the difference between water demand and water supplied for each time period (Equation 12):

$$D_t = \begin{cases} X_t^T - X_t^S, & \text{if } X_t^T > X_t^S \\ 0, & \text{if } X_t^T \leq X_t^S \end{cases} \tag{12}$$

Where $X_t^T$, $X_t^S$ and $D_t$ denote water demand, supplied water, and deficit, respectively, at the time step $t$. The deficit is positive when the water demand is more than what is provided, and it will be zero when the water supply is equal to or greater than the water demand (Loucks, 1997).

Volumetric reliability (Rel) is the total volume of water supply divided by the total water demand (Hashimoto et al., 1982; Vieira and Sandoval-Solis, 2018; Equation 13).

$$Rel = \frac{\sum_{t=1}^{t=n} X_t^S}{\sum_{t=1}^{t=n} X_t^T} \tag{13}$$

Resilience (*Res*) is a measure of the system's capacity to adapt to changing conditions, defined as the probability that the system will remain in a non-failure state (Lane et al., 2015; Moy et al., 1986; Sandoval-Solis et al., 2011; Vieira and Sandoval-Solis, 2018; Equation 14):

$$Res = \frac{\text{No. of times } D_t = 0 \text{ follows } D_t > 0}{\text{No. of times } D_t > 0 \text{ ocurred}} \tag{14}$$

Vulnerability (*Vul*) demonstrates the average severity of a deficit during the total number of months simulated or, in other words, the likely damage from a failure event (Asefa et al., 2014; Kjeldsen and Rosbjerg, 2004; Sandoval-Solis et al., 2011; Equation 15):

$$Vul = \frac{\left( \dfrac{\sum_{t=1}^{t=n} D_t}{\text{No. of times } D_t > 0 \text{ ocurred}} \right)}{\sum_{t=1}^{t=n} X_t^T} \tag{15}$$

The maximum annual deficit (Max. Deficit) is the worst-case annual deficit for the entire period of simulation (Moy et al., 1986). A dimensionless maximum deficit was calculated by dividing the maximum annual deficit by the annual water demand (see Sandoval-Solis et al., 2011; Equation 16):

$$Max.\ Deficit = \frac{\max(D_{annual}^i)}{X_{annual}^T} \tag{16}$$

The sustainability index (SI) is an index that measures the sustainability of reservoir systems and can be used to estimate and compare the sustainability of proposed water policies (Sandoval-Solis et al., 2011). Sandoval-Solis et al. (2011) proposed a variation of Loucks' SI where the index is defined as a geometric average of $M$ performance criteria ($C_m$; Equation 17):

$$SI = \left[\prod_{m=1}^{M} C_m\right]^{\frac{1}{M}} \tag{17}$$

For this research, the proposed sustainability index (SI) for the Folsom Reservoir is formulated in Equation 18.

$$SI = \left[Rel \times Res \times (1-Vul) \times (1-Max.\ Deficit)\right]^{1/4} \tag{18}$$

### Results

This study developed various PGMs to obtain a monthly optimal rule curve over the period of 65 years (1955-2020) for the Folsom Reservoir in California, USA. The performance was tested using various metrics, and results were compared with the base conditions and the SOP under the flood control policy. This section discusses the results of network structure, hyperparameter tuning, the relation between state variables and decision actions, overall comparison of the RL agents, and performance assessment and sustainability.

### Learning Dataset and Network Structure

As mentioned, the primary data set utilized for the learning task of PGMs was 65-years of historical records, including observed hydroclimatic (i.e., streamflow) encompassing a wide variety of extreme hydrologic events experienced in the Sacramento basin and demands at a monthly time step from 1955 (beginning of the Folsom Reservoir operation) through 2020. To enrich the learning data set and to better account for extreme flood and drought periods, an additional 100-years synthetic stochastic streamflow was also produced based on the 65-years of historical records using the synthetic streamflow generator (Qsynth; see Herman et al. 2016). The synthetic stochastic sampling approach better represents the internal variability for the Folsom Reservoir system aiding in overcoming the limitations associated with observation records. Following Herman et al. (2016), monthly synthetic streamflow ensembles were generated using Cholesky decomposition to preserve the autocorrelation of monthly historical records while simultaneously including plausible streamflow values outside their observed range. This method was proved to be successful by several case studies that were implemented by Herman et al., (2016) and Kirsch et al., (2013).

The PGMs' value networks (both main and target) and the soft Q-networks are fully connected neural networks, each with 50 hidden layers, the LReLU activation functions, and an output layer with a linear activation function. The number of neurons in all hidden layers is a hyperparameter to be set by the user. The policy network has two outputs, the mean and the logarithm of the standard deviation (clamped to be in a sane region). These are used for a reparameterization to ensure that the sampling from the policy is differential, and the errors can be appropriately backpropagated, leading to faster convergence. The action taken from a given state is obtained from the policy function by sampling noise from a standard normal distribution, multiplying it with the standard deviation, adding it to the mean, and then transforming it using a sigmoid activation function to ensure the action between 0 and 1.

For all the PGMs developed in this study, the networks made the policy decision updated by performing three general steps (i) initialization of the networks, (ii) initialization of the environment, and (iii) a training loop. At the beginning of each episode for the 1955-2020 period, the environment was reset, and initial reservoir storage was assigned. For every monthly time step, the RL model chose an action from the policy (or randomly in the early exploration phase) and computed the reward and subsequent state variables. In this research, Adaptive Moment Estimation (Adam; Kingma and Ba, 2014) was selected as an optimization algorithm to train all models. Other details such as network architecture were held fixed across conditions. Next, the model saved the obtained vector containing state variables, action, and associated reward to the memory buffer. Networks weights were then updated using the memory buffer. The networks' updating process has the following steps: (i) predicting Q-functions, value, and policy networks for all states in the batch and their suggested actions, (ii) evaluating the policy network to get to the next states, (iii) predicting the target value network, (iv) calculating Q-functions and value network losses and performing one back-propagation (BP; update weights), (v) adjusting the target value function using the next step Q-value, (vi) computing the target value function loss and performing one BP, (vii) calculating the policy network loss and performing one BP, and (viii) updating the target value network. Once the learning episodes were completed, the model weights were saved accordingly.

**Hyperparameters Tuning**

The choice of the network architecture and optimizer, along with the hyperparameters, can have a dramatic effect on the DRL final performance. The hyperparameters control how much the weights of DRL models can be adjusted with respect to the loss gradient. In this study, an agent was identified based on different methods, and the network structures were hardcoded to not consider them as hyperparameters. The agent then took action according to the action policy $\mu(s)$. At the beginning of a learning session, the agent's experience was uniformly distributed in the state spaces (both in the storage and time dimensions). The learning rate at each iteration is then recorded and plotted against loss function. As the learning rate increased, there was a point where the loss stopped decreasing, and instead, it started to increase (see Table 2 for the hyperparameter optimal values). The learning process was conducted for 30,000 episodes (iterations) of monthly 780-time steps. After each time step, the experience $[s, a, r, s']$ of each agent is stored in the memory buffer. The stochastic gradient updates were then applied to the parameters of both actor and critic networks. To track the learning process and understand whether the process was converging, it is decided to observe the mean value of the value function that states how good it is to perform a given action in a given state. Table 2 presents the optimal values for hyperparameters of each method. The total training time, including the function evaluations for each technique, was approximately 25 hours on a Linux High-performance Computing (HPC) machine with 16-Core (2.4 GHz) processors, 256 GB RAM, and 2-Core Nvidia P100 GPU processors. Generally, the models converged, although some level of noise was always present. The results array was stored after each evaluation, and the best array among the last ten iterations was selected to get the output of the learning process.

Table 2. The optimal values of PGMs hyperparameters identified based on a trial-and-error process.

| Model | Gamma | Tau | Buffer Size | Critic Learning rate | Actor Learning rate | Batch size | Q learning rate | Alpha |
|---|---|---|---|---|---|---|---|---|
| DDPG | 0.99 | 0.01 | 1e6 | 1e-4 | 1e-3 | 64 | - | - |
| TD3 | 0.99 | 0.01 | 1e6 | 1e-4 | 1e-3 | 64 | - | - |
| SAC18 | 0.99 | 0.01 | 1e6 | 1e-3 | 3e-3 | 64 | 3e-3 | 0.2 |
| SAC19 | 0.99 | 0.01 | 1e6 | 1e-3 | 3e-3 | 64 | 3e-3 | 0.2 |

As mentioned before, Adam optimizer was selected as an optimization algorithm to train all models. Adam combines the advantages of two SGD extensions (RMSProp; Root Mean Square Propagation, and AdaGrad; Adaptive Gradient Algorithm) and computes individual adaptive learning rates for different parameters. In order to fairly compare the results across multiple methods, all architectural details that are not specific to the method under consideration were held fixed across conditions. Analysis revealed that a lower learning rate for the policy network than the value functions (actor network learning rate = 0.1 critic network learning rate) is more robust to collecting experience at a faster rate than the policy adapts to the experience. A lower learning rate was able to control and adjust the weights of the network with respect to the loss gradient.

**Relationship between State Variables and Decision Action**

All the operating policies designed by the DDPG, TD3, SAC18, and SAC19 were simulated under historical conditions (measured reservoir inflows, evapotranspiration rates, and the Folsom Reservoir properties). Figure 7-a synthetically illustrated the state variables along with the resulting actions of the four PGMs employed in this study. In all cases, the agents were trained for 30,000 episodes corresponding to 65 years of monthly training data. The agents successfully learned a meaningful policy for reservoir operation policy, given the limited state information. When the state variables associated with storage (depending on the inflow value) and the time are dominated by fluctuations, the resulting actions partially mimic the states and show a range of fluctuations in all methods. Considering the policy actions provided in Figure 7-a, the TD3 and SAC18 mostly presented the higher and the lower extreme policy actions, respectively, while the SAC19 and DDPG provided moderate decision actions.

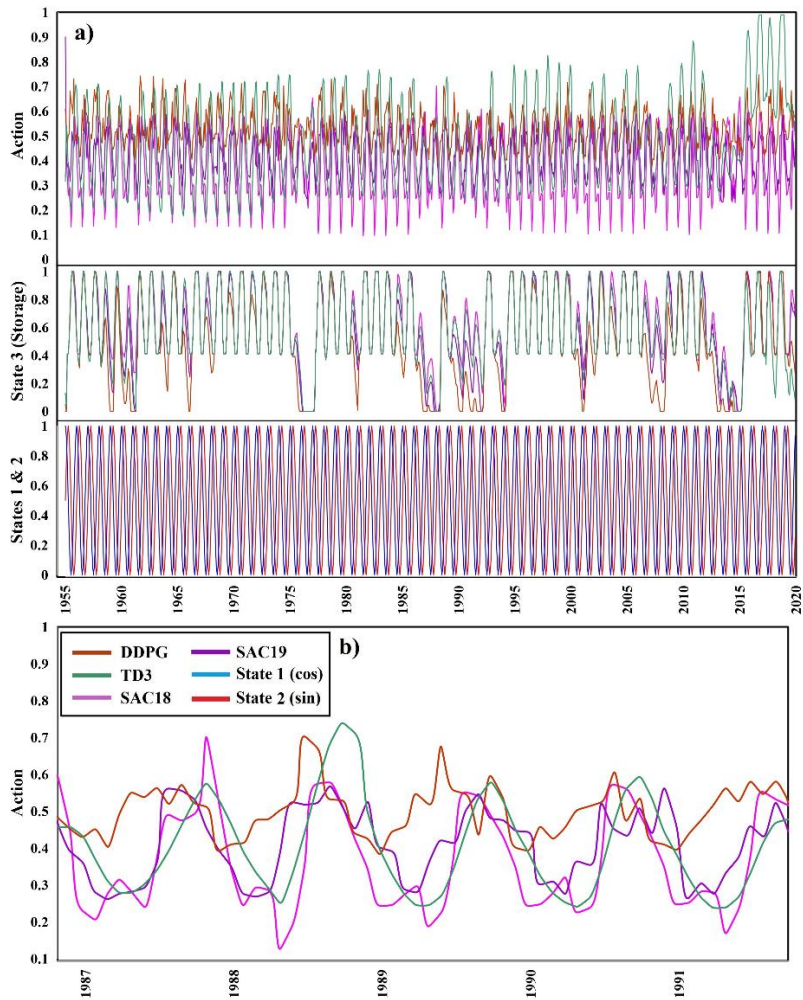

Figure 7. a) The relationship between the state variables and optimal policy actions taken by the PGMs. b) Optimal policy actions during the dry period (1987-1992) for the Folsom Reservoir.

Concerning the optimal policy actions provided by each method, as expected, the DDPG method overestimated the value function and revealed a lower range of variability than its variants. On the other hand, the TD3 illustrated a smoother sequence of decision policy action with a higher variability range, demonstrating the model's ability to react to different conditions quickly. It is interesting to note that there are more fluctuations in the actions taken by both SAC18 and SAC19 methods than the DDPG and TD3. This behavior is due to entropy optimization, where the SAC agent automatically optimizes the entropy and increases the random exploration for regions of the parameter space with large fluctuations. This behavior is part of the reason why SAC is ideal for stochastic-dominated environments where the actor performs the same action and fails miserably when the condition changes. Figure 7-b presents the policy actions for a 5-year dry period (1987-1992) that illustrated the difference among models' optimal policies when the reservoir inflow significantly decreased over time. A 5-year dry period was identified based on a 5-year monthly moving average of reservoir inflow over 65 years of the simulation period.

**Optimal Operation Policy Actions**

Figure 8 presents the rewards associated with the optimal policy actions provided by each method. As mentioned before, the reward (or penalty) in each time step originated from the supply water deficiency and generated power amounts based on Equation 8. The agent's goal is to maximize the cumulative rewards and learn by adjusting its policy (the agent's strategy) based on the obtained rewards. The results for the entire simulation period showed both SAC18 and SAC19 with cumulative rewards equal to -426409 and -426077, respectively (see Table 3), outperforming the TD3 and DDPG results. This can also be realized by looking into reward values and their components presented in Figure 8.

Table 3. Performance criteria results for the PGMs as well as base conditions and SOP. Bold values represent the best performances.

| Method | Reliability (Volume) | Resilience | Vulnerability | Max Annual Deficit | SI | Ave. Annual Power Production (GWh) | Cum. Rewards |
|---|---|---|---|---|---|---|---|
| DDPG | 0.91 | 0.39 | 5.18E-04 | 0.76 | 0.54 | 683.60 | -556289 |
| TD3 | 0.91 | 0.37 | 4.52E-04 | **0.62** | 0.60 | 705.86 | -459503 |
| SAC18 | 0.91 | **0.45** | 4.35E-04 | 0.63 | **0.62** | **708.76** | -426409 |
| SAC19 | 0.91 | 0.38 | **3.74E-04** | 0.66 | 0.59 | 701.52 | **-426077** |
| SOP | **0.97** | 0.23 | 8.09E-04 | 0.71 | 0.50 | 700.46 | - |
| Baseline | 0.90 | 0.27 | 3.96E-04 | 0.70 | 0.56 | 691.35 | - |

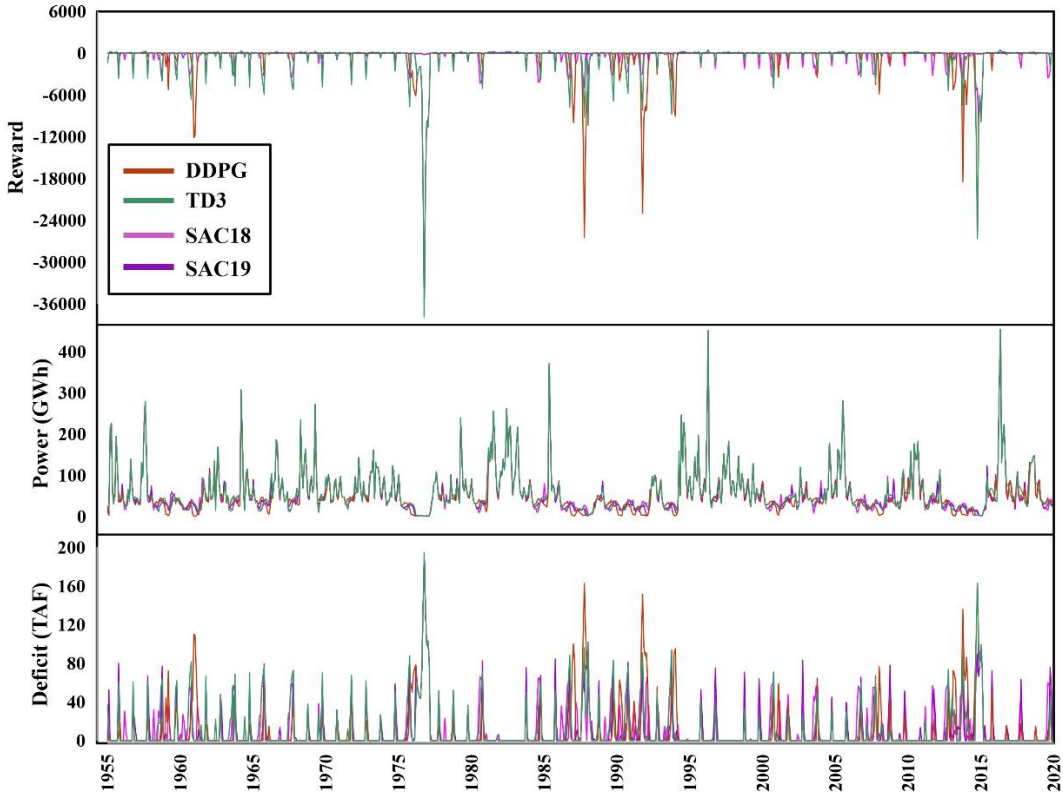

Figure 8. Reward and its components as a result of policy actions taken by each method.

Figure 9 illustrates the monthly release values optimized by each policy action method as well as the SOP and historical releases (base conditions). The monthly release values during a 5-year dry period revealed that the DRLs' performance was more reliable than the base conditions with respect to demand satisfaction. As noted, during wet years, the PGMs agents increased release amounts in anticipation of high inflow, which led to the lowest water level and also reduced the volume of spilled water over the year (see Figure 9). Contradictory during dry years, the PGMs' agents released less water to maintain high water levels in anticipation of low flow conditions ahead of time. Figures 10-a and b illustrated the Folsom monthly storage levels during the simulation period as well as during a 5-year dry period to evaluate the state of the Folsom Reservoir based on different policies identified by the PGMs, SOP, and base conditions.

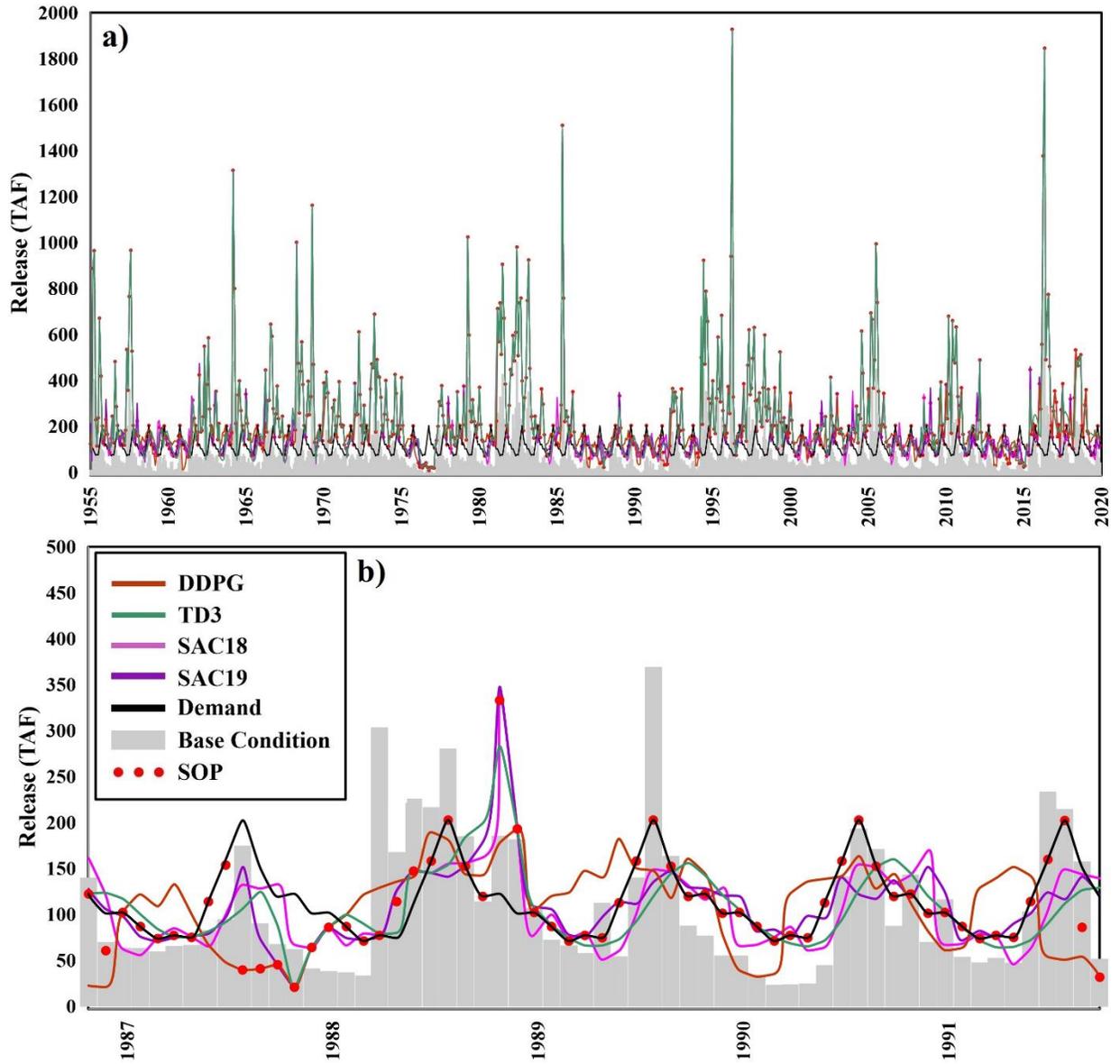

Figure 9. a) Monthly release values obtained from optimal policy actions provided by each PGMs. b) Monthly release values during a 5-year dry period (1987-1992).

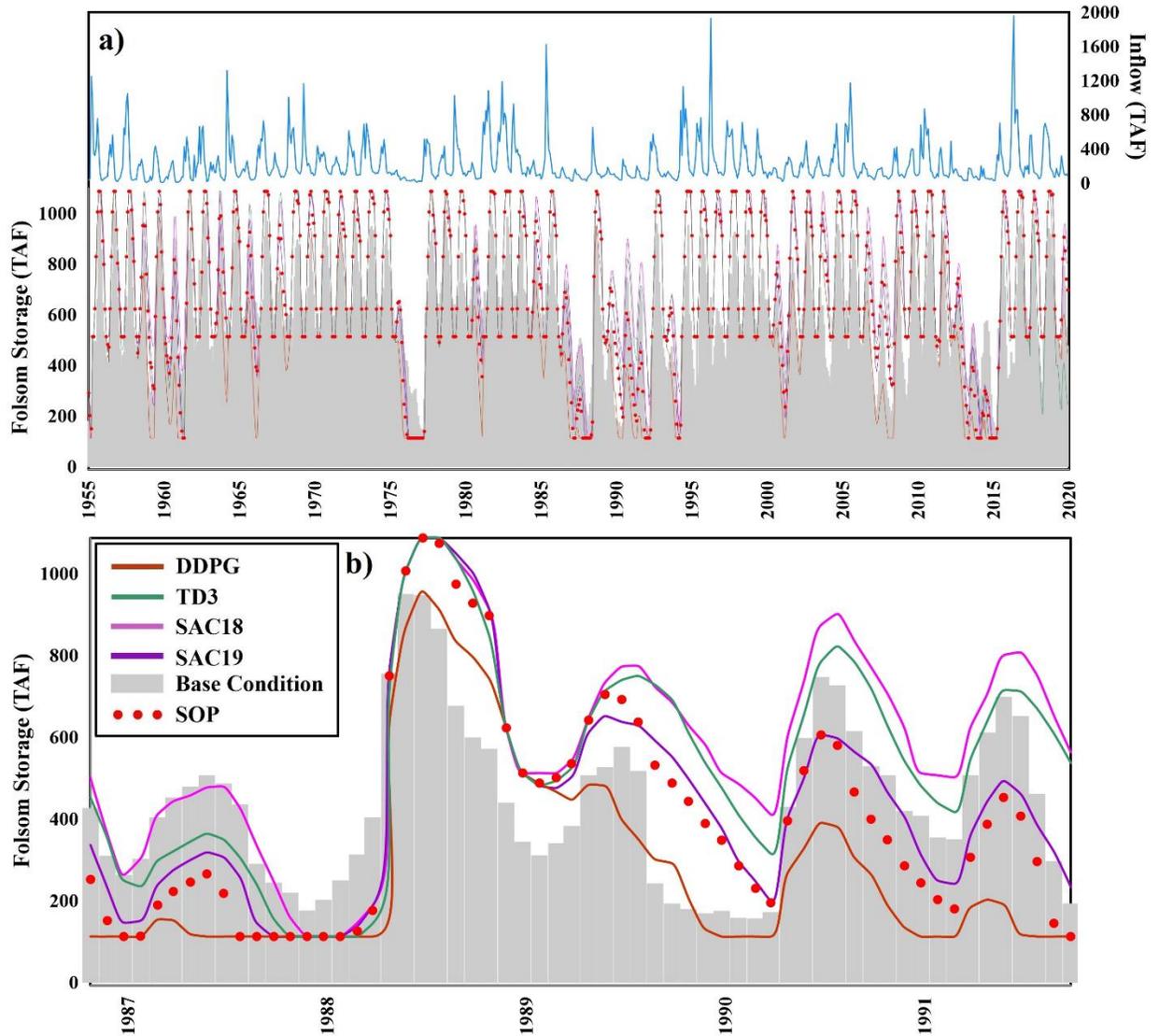

Figure 10. a) Folsom monthly storage amounts identified by the PGMs, SOP, and the base conditions during the simulation period. b) Reservoir storage levels during a 5-year dry period.

Results indicated that the TD3 and both SAC18 and SAC19 predicted a higher reservoir storage level than the SOP and base conditions during the dry period. As discussed before, SOP releases water as close to the delivery target as possible, saving only surplus water for future delivery. This is because SOP is strict and more practical during periods of operation when inflow is plentiful. However, it neglects to consider potential shortage vulnerability during later periods, and strict obedience to this rule is practically rare due to a desire to maintain the reservoir average level and avoid extremely severe shortages.

To evaluate the reliability of different methods in the case of supplying demands, Figure 11 presents the annual deficit (%) calculated based on the optimal decision actions provided by each method during the simulation period. As illustrated, all PGMs excluding the DDPG successfully minimized the deficiency and performed relatively well compared to the base condition. As expected, the maximum deficiency in supplying water demands occurred during the dry period (1987-1992).

Figure 12 illustrates PGMs' performances in generating power that is totally dependent on the reservoir state variable (head of water over turbine). As illustrated, the PGMs' agents learned to release the majority of water when there is a high level of water over the turbine to maximize hydropower generation. This implies the fact that DRL is flexible with learning the objectives and provided various operation policies to maximize the total hydropower benefit in response to dynamic inflow conditions. In contrast, SOP generated releases with less variation and could not adjust outflows due to its routine operation procedure.

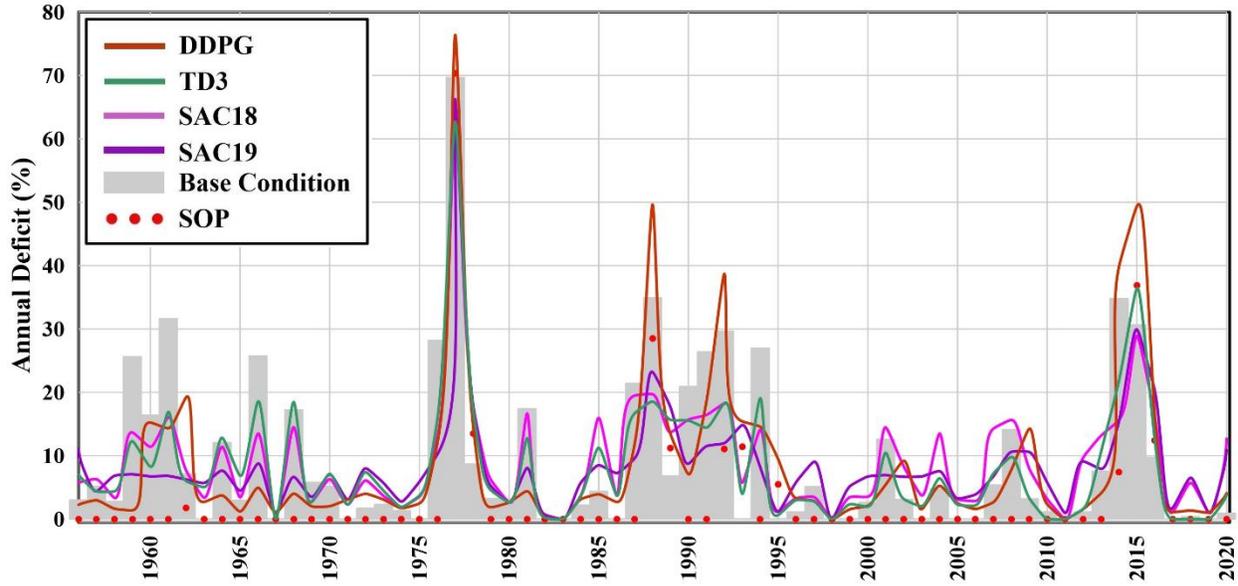

Figure 11. Annual deficit (%) calculated based on the optimal decision actions provided by each method vs. SOP and base conditions.

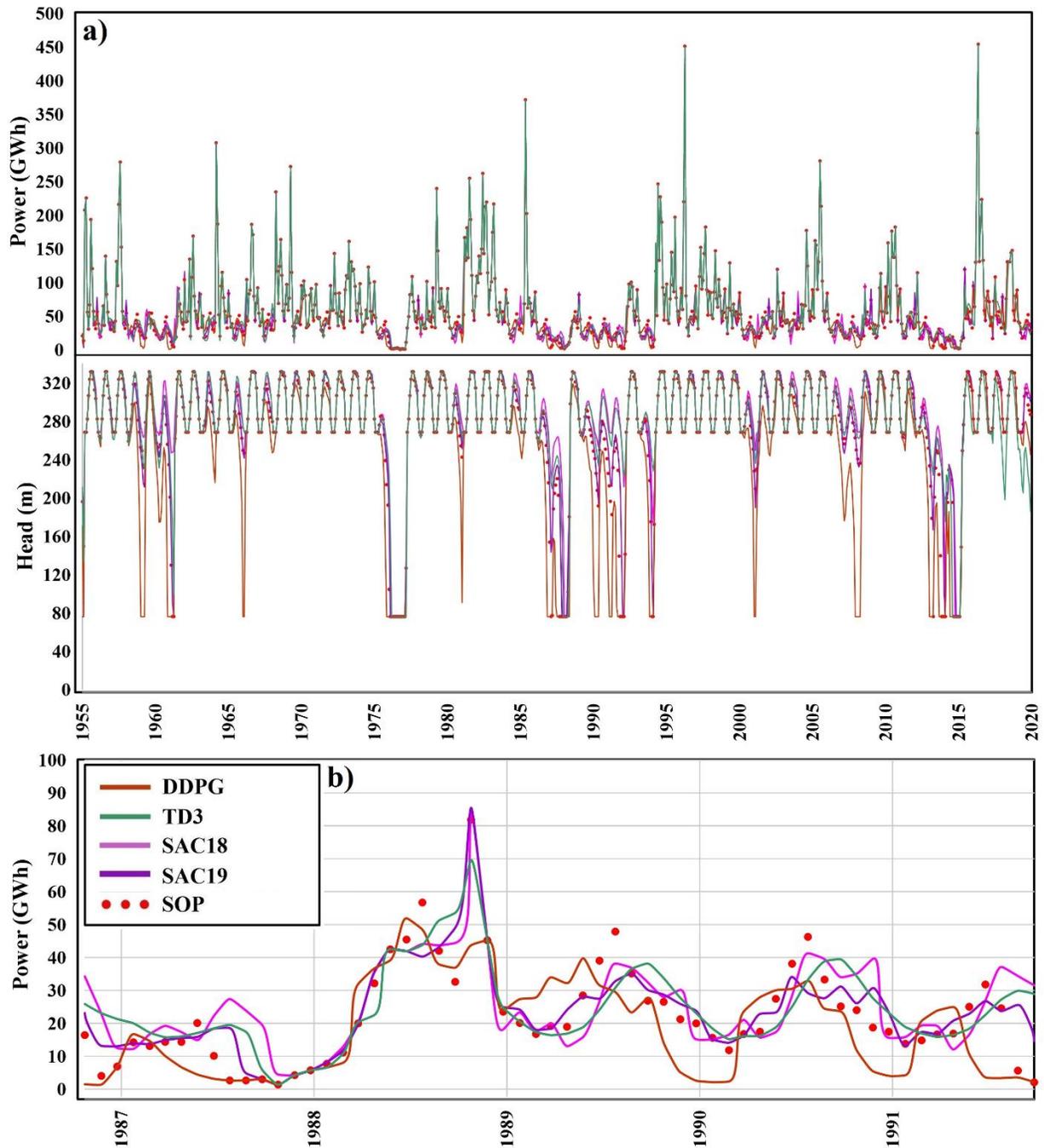

Figure 12. a) Generated power based on the suggested operating policy by each method as well as related reservoir state variables (head of water over the turbine). b) Generated power over dry period (1987-1992).

**Performance Assessment and Sustainability**

Table 3 presents the performance criteria such as reliability, resilience, vulnerability, and maximum deficit, as well as the annual average, generated power for the DDPG, TD3, SAC18, SAC19, SOP, and base conditions. The cumulative rewards over the simulation period are also provided for the DRL methods.

In terms of different sustainability factors, SOP, due to its policy's nature, delivered water as close as to the target demand to show more reliable (volumetric) results in supplying water demand than the other methods. Overall, all PGMs showed similar results to the base conditions in the case of volumetric reliability of supplying demands. However, the PGMs presented promising results in the case of resilience and robustness, which reveal the ability of the system to recover and bounce back from a failure compared to the SOP and base conditions. If failures are long-lasting events and system recovery is slow, this may cause serious consequences. Among multiple PGMs used in this study, SAC18 showed the most promising results in terms of robustness and sustainability. In addition, the vulnerability index is computed to see how severe the consequences of a failure in different policies could be. As expected, the operating policy suggested by SOP was the most vulnerable among all methods as its operation was along with severe deficiency. In contrast, SAC19 with the minimum vulnerability index shows more acceptable failure consequences against severe conditions and dry periods. With respect to different PGMs actions policies, DDPG, with the maximum annual deficit of around 76 percent, showed quite poor performance even after a quite large number of iterations (30,000). TD3, on the other hand, maintained the lowest maximum annual deficit (~62%).

Combining all four aforementioned indices into an SI, the SAC18 method provided the best policy in terms of sustainability performance. In addition, in terms of generating hydropower, SAC18 provided a more reliable policy compared to the other methods. However, considering the base conditions in generating power (average annual production of 691 GWh based on the USBR Environmental Water Account report, 2012), all the PGMs policies except DDPG showed promising results in providing an optimal policy for maximizing power production. From an optimization problem point of view, compared to other methods, SAC19 was found to be a more intelligent and reliable policy method for the Folsom Reservoir sequential operation policy decisions with respect to the maximum value of the objective function (Cum. Rewards). However, the SAC18 results were also promising, with slightly less objective function value compared to the SAC19. Both SAC18 and SAC19 were robust against a range of perturbations in the environment. Visualizing the learned policy, it is observed that the SAC18 and SAC19 policies used many different routes to satisfy the goal function. A central feature of both SAC18 and SAC19 is entropy regularization. These policies are trained to maximize a trade-off between expected return and entropy (randomness in the policy). This has a close connection to the exploration-exploitation trade-off: increasing entropy results in more exploration, which can accelerate the learning process later on. It can also prevent the policy from prematurely converging to an inappropriate optimum value.

**Conclusions and Future Work**

This study presents a generalized framework for creating an intelligent agent for reservoir operation policy decision and control leveraging DRL methodologies. While DRL has been used successfully in the computer science communities, to our knowledge, our work is the first attempt to formally characterize the robustness of multiple DRL policy gradient algorithms for the control of a reservoir operating system, even though numerous studies have conjectured that such robustness results may be possible. Different variants of PGMs such as DDPG, TD3, SAC18, and SAC19, were applied to solve the DP problem and tackle dimensionality issues for the Folsom Reservoir operation policy decisions in California, USA. The results were then compared with the SOP and baseline conditions using different performance criteria and sustainability metrics. The proposed PGMs relied on continuous action spaces to sample actions from an appropriately parameterized Gaussian distribution. The agents can interact with any simulation model of the environment, utilize any user-defined reward function, and evaluate a wide array of options and variations for PGMs optimization.

Our results suggest that stochastic, entropy maximizing DRL algorithms can provide a promising avenue for improved robustness and stability and further exploration of reservoir operation policy decision problems. Overall, the use of historical records and synthetic stochastic datasets for agent learning indicated that the DRL approach as an alternative to the tabular approach, along with the agents' design as presented, could adapt to stochastic conditions and function approximation through the use of DNN. Results showed that among different PGMs, both versions of the SAC (SAC18 and SAC19) are competitive and are capable of outperforming other policy methods across high-dimensional control tasks. Indeed, SAC18 effectively produced long-term strategies for the use of the stored water, focusing on mitigating water supply deficiency. This best policy approach can be used as a decision-support framework for reservoir operation considering hydropower production, flood management and control, and water supply during drought periods. During droughts, California relies on water stored in surface reservoirs to help offset shortfalls in precipitation. These key policy decisions could be made in advance of droughts when there is time for meaningful stakeholder engagement and for considering the implications of different policy options.

Given the recent emergence and popularity of DRL, much research is still needed to understand its full potential as a viable technique/tool for reservoir operation management. Our research reveals a number of advantages and also challenges related to the DRL application. The ability of DRL to simply pass the learning task and optimize the policies without any concerns about complexities, non-linearities, and formulations that sometimes hinder other control approaches, appears to be the major advantage of utilizing this technique for reservoir operation. However, there are a number of important caveats, such as the challenges in selecting function approximators, determining the complexity of reservoir control problems, and dealing with real-world implementation concerns. Contrary to the framework and learning function formulations, the implementation may be hampered by the inherent complexity and objectives of control problems.

In this research, an Actor-Critic learning procedure is implemented which is most sensibly used in a (quasi) stationary setting with a continuous state and action space. However, the choice of the right features for the actor and critic remains an open research question. The policy gradient theorem recommended that it is best to first choose a parameterization for the actor, after which compatible features for the critic can be derived (Grondman et al., 2012). In this regard, the use of compatible features for the actor and critic can lessen the burden of choosing a separate parameterization for the value function. To reduce the variance in the policy gradient estimates, state-dependent features should be included in the value function in addition to compatible features. Although, how to choose these additional features remains a difficult task. In addition, balancing the exploration and exploitation of a policy and choosing good learning rate schedules are critical for improving the speed of learning. In this sense, natural gradient actor-critic methods are still a very promising area for future research as it speeds up the learning without any real additional computational effort.

This study demonstrated that a DRL agent can efficiently manage a single reservoir system although simultaneously handling multiple reservoirs may be challenging. The rise in the number of states and actions that must be represented for multiple reservoirs using the NNs is one of the primary challenges. While more computational time may alleviate this issue, the NN structure may also need to be modified. In a cascade reservoir system scenario, actions taken at one reservoir may impact another reservoir later on. As a result, the agent would benefit from a planning-based approach that took into account both current and future states. Such planning-based strategies have been developed in the RL literature and should be examined to realize if they can enhance policy decisions and modeling performance (Clavera et al., 2018; Depeweg et al., 2017; Hamilton et al., 2022). Finally, an unresolved challenge affecting most reservoir operation policy methods is their limited uptake by practitioners (Giuliani et al., 2021). The black box perception of system models hinders the uptake of DRL research results which makes their recommendations difficult to understand and justify (Castelvecchi, 2016). Perhaps creating local champions throughout all phases of the modeling process to voice value stakeholders' inputs, is an important step moving forward for water resources system management (see Reed and Kasprzyk, 2009; Fletcher et al., 2022).

There are several promising avenues for future work. An interesting direction would be to use the proposal distribution for more intelligent action exploration than random exploration. Efficient exploration with a meta-controller policy (Finn et al., 2017) that adapts the exploration to different objectives and the long versus short-term behavior of the agent could be particularly successful. Meta-learning helps yield dataset-specific priors for hyperparameter optimization. To make the DDPG algorithm a meta-DRL algorithm, a batch of prior task-specific experiences can be fed into an embedding network which produces a low-dimensional latent variable (see Rakelly et al., 2019). These recent advances can improve reservoir operation policy by optimizing the model parameters for fast adaptation and learning an optimal control law. In addition, the DNN structure can be updated to take temporal aspects into account (e.g., recurrent neural networks; RNNs) or structured information in the form of graphs. Maximum entropy methods such as those that incorporate second-order information (e.g., trust regions; see Schulman et al., 2015b) or more expressive policy classes can also be an exciting avenue for future research. In the case of cascading reservoirs, further constraints can be considered in the framework which can be defined through OpenAI Gym. The problem can then be solved by employing multiple agents associated with different reservoirs to discover optimal policy actions. It is also important to note that, due to the increased size of the action space, it is expected that the model runtime take longer, and the data must be sufficient for the training.

Although our research lays the foundation for the DRL algorithms application with the idea of finding an optimal continuous control policy for reservoir operation management and control, the framework presented here is flexible, and the DRL methods can be easily applied to different time scales or extended with additional constraints, e.g., cascading reservoirs, minimum power production requirements, and non-linear production functions. As a whole, this research investigates how DRL approaches can be implemented in the evaluation and planning of reservoir operation policy within complex water supply networks. The outcomes of this research hold significant promise for better addressing of reservoir policy operation challenges in California and may provide a road map for parameterization of other decision policy problems in other regions.

**Declaration of Competing Interest**

The authors declare that they have no known competing financial interests or personal relationships that could have influenced the work reported in this paper.

**Data availability statement**

All data, models, or codes that support the findings of this study are available from the corresponding author upon request.

**Acknowledgments**

This research is supported by the US Geological Survey (grant number # 5001-20-207-0312-216-2024917). Clemson University is acknowledged for its generous allotment of computing time on the Palmetto cluster. The authors would like to thank M. Giuliani and A. Castelletti of Politecnico di Milano, Milano, Italy for their constructive comments on the methodology and synthetic streamflow generator approach.

**Supplementary Information**

**Introductory to Reinforcement Learning**
The nature of reservoir operation problem led to the use of RL as a means of estimating the state-value function. RL refers to a class of machine learning (ML) methods in which an intelligent agent can learn advantageous behaviors from interactions with its environment. The intelligent agent is given an optimization goal to conduct actions that, along with possible random input, result in changes to the environment. A reward signal from the environment alerting the agent of the value of the action taken, as well as a state signal returning the environment's consequent state.

**Markov Decision Process**
The Markov Decision Process (MDP) serves as the foundation for the design of an RL problem. A finite MDP problem consists of a discretized state, $S_t \in S$, a set of actions that can be taken in a given state, $A_t \in A(s)$, and the rewards that can be achieved, $R_t \in R$. Through the system dynamics equation, the received reward and the system's resulting state are associated with the initial state and chosen action. After starting in state $s$ and performing action $a$, equation S1 gives the probability of finding the system in state $s'$ and getting a reward $r$.

$$p(s', r \mid s, a) \doteq \Pr\{S_t = s', R_t = r \mid S_{t-1} = s, A_{t-1} = a\} \qquad \text{S1}$$

The agent maximizes expected cumulative rewards over a sequence of time steps as follows:

$$G_t = R_{t+1} + R_{t+2} + \cdots + R_{t+T+1} \qquad \text{S2}$$

The concept of discounted returns is introduced to avoid infinite returns and to reflect the diminished value given to future rewards.

$$G_t = \sum_{\tau=0}^{T} \gamma^\tau R_{t+\tau+1} \qquad \text{S3}$$

where $\gamma \in [0,1]$ and $T$ is the number of time steps. As a result, the goal is to determine a set of actions that maximize the expected value of the return (Equation S4).

$$J = \mathrm{E}\left[G_t \mid S_t = s\right] \qquad \text{S4}$$

**Value Functions and Optimal Policies**
DP and other methods for solving MDPs often consider two value functions. The state-value function, $\upsilon_\pi(s)$, reflects the return's expected value based on a known state and actions taken according to a policy (Equation S5). The policy is allowed to be stochastic or deterministic. In the stochastic case, $\pi(a \mid s)$ is a mapping from a state to the probability of taking action $a$. In the deterministic case, the policy maps from a state to an action and the notation is adjusted to $\mu(s)$ in order to indicate the difference. The action-value function ($q_\pi$) describes the expected reward for a given state conditioned by taking a particular action and following the policy $\pi$ at all subsequent stages (Equation S6). The Bellman equations, a recursive characteristic of value functions, are used to solve this problem (Bellman, 1957). Equation S7 presents the state-value function based on the Bellman equation.

$$\upsilon_\pi(s) \doteq E_\pi\left[G_t \mid S_t = s\right] \qquad \text{S5}$$

$$q_\pi \doteq E_\pi \left[ G_t \middle| S_t = s, A_t = a \right] \tag{S6}$$

$$\upsilon_\pi(s) = E_\pi \left[ R_{t+1} + \gamma \upsilon_\pi(s') \middle| S_t = s \right] \tag{S7}$$

Therefore, the agent can find the optimal policy $\pi^*$ based on state value function (Equation S8). For the optimal policy and all possible states, Equation S9 presents the action-value function. Thus, the state-value and action-value functions are related by Equation S10.

$$\upsilon^*(s) = \max_\pi \upsilon_\pi(s) \tag{S8}$$

$$q^*(s,a) = \max_\pi q_\pi(s,a) \tag{S9}$$

$$\upsilon(s) = \max_a q(s,a) \tag{S10}$$

Combining optimality notation with the recursive characteristic of the value functions, the agent can treat the problem as a greedy optimization of the return over the next stage plus the value of the resulting state, which is represented by the Bellman optimality equations:

$$\upsilon^*(s) = \max_a E \left[ R_{t+1} + \gamma \upsilon^*(S_{t+1}) \middle| S_t = s, A_t = a \right] \tag{S11}$$

$$q^*(s,a) = E \left[ R_{t+1} + \gamma \max_{a'} q^*(S_{t+1}, a') \middle| S_t = s, A_t = a \right] = E \left[ R_{t+1} + \gamma \upsilon^*(s') \right] \tag{S12}$$

The optimal value functions (Equations S11 and S12) can be computed using SDP by assuming an initial state value, iteratively solving backward the Bellman optimality equations, and updating the current value at each iteration (see Soncini-Sessa et al., 2007). This process is repeated until a satisfactory level of convergence is achieved.

The agent's current view of the state provides all information required to characterize the environment, which is implicitly identified in the problem formulation. To facilitate, the current state should have all of the information required to provide reliable probabilities to the state dynamics equation. A partially observable state occurs when there is insufficient information about the state of the system, and the problem becomes a Partially Observable MDP (POMDP). Many studies have been conducted to solve the POMDPs problem such as Lovejoy, (1991), Jaulmes et al., (2005), and Ross et al., (2008), among others.

As many elements contribute to the transition from one state to the next, such as historical precipitation and soil moisture, which are not always available, a realistic assessment of a reservoir operation problem may require considering the state as partially observable. Besides the issue of the partially observable state, there is the issue of relying on probability distributions to express the state dynamics equation. The transition probabilities for a reservoir system can be imprecise and the distribution might be unable to capture the temporal and spatial correlations of the random variables when there are more than a few states or random variables, or when time steps are short. Furthermore, when a system is evaluated at a fine temporal resolution, routing effects may lead to rewards that are realized several time steps after the decision was made, a concept known as delayed rewards. This is an issue that a probabilistic model cannot address it appropriately.

**Temporal-Difference Learning**

RL and DP have a lot in common, however utilizing trial and error approaches such as Monte Carlo (MC) is one of their dissimilarities (Sutton and Barto, 2018). The process of learning in RL relies on interactions between environment and reinforcing information (identified as rewards) and it does not require any knowledge of a probabilistic model of the environment. Methods that don't make use of probabilistic models are so-called model-free procedures. Temporal-difference learning, which is utilized in the highly successful Q-learning approach, is another feature of some RL methods (Watkins et. al., 1992). In this method, the agent stores an estimate, $Q(s,a)$, of the $q(s,a)$ function and updates the estimate by iteratively interacting with the environment and getting a reward at each time step. The agent takes actions based on a policy that insures exploration such as the ò-greedy method, in which for a value of $ò \in [0,1]$ the action $A_t = \underset{a}{argmax}\, Q(S_t, a)$ is selected with probability $1-ò$, and a random action is selected with probability ò. The estimate updates as follows:

$$Q(S,A) \leftarrow Q(S,A) + \alpha \left[ R + \gamma \max_a \{Q(S',a)\} - Q(S,A) \right] \qquad \text{S13}$$

The update enhances the current estimate, $Q(S,A)$, in the direction of a new estimate by a step size $\alpha$. The temporal difference error (TD-error), the difference between the previous estimate, $Q(S,A)$, and the new estimate made up of a single step reward and the best possible value of $Q$ at the next state, $R_t + \gamma max_a Q(S',a)$, is defined as follows:

$$\delta_t = R + \gamma \max_a Q(S',a) - Q(S,A) \qquad \text{S14}$$

The method is referred to as an off-policy method since the new estimate is based on the possibility of taking the action that maximizes $Q$ rather than the action described by a fixed policy $\pi$.

The $Q$ function becomes a look up table for discrete action and state spaces, and thus is referred to as a tabular technique. In the tabular case, Q-learning has been proved to be converged to an optimal state-value function (Watkins et. al., 1992). The convergence proof requires visiting all states across an infinite number of times, which is problematic to implement, hence methods like ò-greedy action selection is used to ensure state-space exploration.

### Value Function Approximation

RL in continuous state spaces requires function approximation to represent a state action with a parameterized function. Value function approximation provides a more compact representation that generalizes across states or states and actions. There are many possible function approximators including linear combinations of features, neural networks, decision trees, nearest neighbors, etc. This research used DNNs as an approximator of the value functions. There are various possible configurations to employ DNNs for approximating value functions. In this study, only fully connected feed-forward neural networks (FFNNs) were employed. The state vector is identified in the network's first layer as an input, whereas the final layer presents the network's output, and the prior layers are referred to as hidden layers. Depending on the application, there are various forms of activation functions to map the transformed input to a single value output. In this study, the leaky rectified linear unit (LReLU; also known as the parameterized rectified linear unit) was selected as it is easy to differentiate, does not suffer from vanishing gradients problem, and the leaky characteristic of the LReLU solves the problem of dying nodes (the dying ReLU refers to the problem when ReLU neurons become inactive and only output 0 for any input; see He et al., 2015). The LReLU activation function is used for all hidden layers, and a linear function is selected for the output layer as an activation function that simply passes along the matrix transformation and allows the output to take on any real number. In addition, in the case of the action policy network output function, the sigmoid and hyperbolic tangent functions were used to ensure the bounds of the maximum/minimum release constraints.

### RL Methods with Continuous Action Spaces

As previously stated, the tabular RL approaches suffer from the curse of dimensionality as the number of state and action variables grows, and the state and action spaces become more finely discretized. To overcome this issue, the Q function should be estimated using a function such as artificial neural networks that can serve as an approximator. DNNs are considered universal function approximators, however different studies demonstrated that employing non-linear functions as approximators shows instability (Tsitsiklis and Roy, 1997; Sutton and Barto, 2018). Mnih et al. (2015) proposed a replay buffer and incremental updates to overcome instability issues both of which reduce the impact of correlation between samples (Mnih et al., 2015). However, this method again can only be applied to discretized actions and suffers from the same dimensionality problem as the action space grows. There is also an embedded optimization step with the Q-learning update, which can be time-consuming. Lillicrap et al. (2015) developed the DDPG technique, which is able to use a continuous action space by coupling the deterministic policy gradient (DPG) method with an advanced version of the deep Q network (DQN; Lillicrap et al., 2015).

**Experience Replay Buffer**

Mnih et al. (2015) proposed an experience to reply to stabilize the RL process in the case of using DNNs as value function approximators. In this strategy, the agent's experiences will be memorized in a buffer, $D$, with a size of $D_{max}$. A single experience vector includes the current state, the selected action, the subsequent reward, and the state transitions that occur as a result of the performed action (tuple of $[S, A, R, S']$). During DNN training of this research, mini batches of memories were randomly taken from the buffer and utilized to update decision action. The oldest memories were diminished, and new experiences were stored in the buffer once the buffer's maximum capacity was reached.

**Policy Gradient Methods**

Deep RL (DRL) is a new type of RL that has emerged as a result of advances in deep learning (DL) techniques. DRL can deal with high-dimensional inputs such as photos, recognize complicated patterns, and extract their features (Arulkumaran et al., 2017). Policy gradient methods (PGMs) are a class of DRL approaches that use gradient descent (GD) to optimize parametrized policies in terms of expected return (long-term cumulative rewards). They overcome many problems plaguing traditional RL approaches, such as the lack of value function guarantees, the intractability problem deriving from uncertain state information, and the complexity arising from continuous states and actions.

All PGMs employed in this study are based on actor-critic networks which is a temporal difference (TD) version of policy gradient. Although the actor and critic networks are separate NNs, they are connected and collaborate constantly. An actor performs the action, while a critic evaluates the actor's performance. Based on the critic's gradient, the actor conducts actions in the environment. The critic collects information about the environment and assigns a reward value to the actor's proposed action. Generally, the actor and critic interactions determines how agent learns, which is based on a reward function that reinforces the learning (Silver et al., 2018). The most relevant PGMs employed in the current study are discussed below (mostly adapted from Achiam, 2018).

**Deep Deterministic Policy Gradient**

Here, the math behind the two parts of DDPG (DPG and DQN) is presented as follows. Learning an approximator begins with the Bellman equation. Equation S15 shows the optimal action-value function described through the Bellman equation that is given by:

$$Q^*(s,a) = \underset{s' \sim P}{\mathrm{E}}\left[ r(s,a) + \gamma \max_{a'} Q^*(s',a') \right] \qquad \text{S15}$$

Where $s' \sim P$ denotes the new state, $s'$, is sampled from distribution $P(\cdot | s, a)$ by the environment. Supposing the approximator $Q_\phi(s,a)$ as an NN with parameters $\phi$ and a set of transitions $D = (s, a, r, s', d)$, Equation S16 presents the mean-squared error (MSE) of the Bellman function.

$$L(\phi, D) = \underset{(s,a,r,s',d) \sim D}{E}\left[\left(Q_\phi(s,a) - \left(r + \gamma(1-d)\max_{a'} Q_\phi(s',a')\right)\right)^2\right] \quad \text{S16}$$

Where $d$ is a binary variable indicating that if the state $s'$ is terminal.

In DQN computation, the aim is to minimize Equation S16, which is a loss function that shows how close $Q_\phi$ comes to satisfying the Bellman equation. The loss function shows the error between the Q-function and the term target (Equation S17).

$$r + \gamma(1-d)\max_{a'} Q_\phi(s',a') \quad \text{S17}$$

DQN and DDPG employed two tricks in the process of minimizing the mentioned loss function including (i) replay buffer and (ii) target networks. As previously mentioned, the replay buffer is a large enough space containing a wide range of experiences that helps DDPG to have a stable behavior.

The process of minimizing loss would be unstable as the target term relies on the same parameters that needed to be trained. The solution to this problem is to employ a second network called the target network, $\phi_{targ}$, which lags behind the first. Thus, a set of parameters for the target network can be used to reach approximately $\phi$ with a time delay. In the DDPG algorithm, the target network is updated using Polyak averaging method (PAM) once per the main network gets updated (Equation S18).

$$\phi_{targ} \leftarrow \rho\phi_{targ} + (1-\rho)\phi \quad \text{S18}$$

where $\rho$ is Polyak hyperparameter ranging between 0 and 1 that needs to be optimized.

In the continuous action spaces, computing the maximum over actions in the target is challenging. To deal with this, DDPG uses a target policy network (similar to the target Q-function) to compute an action that approximately minimizes $Q_{\phi_{targ}}$. The resulting Q-learning with the stochastic gradient descent (SGD) method is performed by minimizing the following loss function.

$$L(\phi, D) = \underset{(s,a,r,s',d) \sim D}{E}\left[\left(Q_\phi(s,a) - \left(r + \gamma(1-d)Q_{\phi_{targ}}(s', \mu_{\theta_{targ}}(s'))\right)\right)^2\right] \quad \text{S19}$$

where $\mu_{\theta_{targ}}$ is the target policy.

In the DDPG policy learning process, a deterministic policy, $\mu_\theta(s)$, should be learned to take actions that maximizes $Q_\phi(s,a)$. In the continuous action spaces, only a GDM can be used to solve Equation S20, assuming the Q-function is differentiable with respect to action.

$$\max_\theta \underset{s \sim D}{E}\left[Q_\phi(s, \mu_\theta(s))\right] \quad \text{S20}$$

To improve DDPG policy exploration, recent studies suggested adding a time-uncorrelated Gaussian noise to the actions during the training. However, to determine how well the policy exploits the learned information, it doesn't need to be added during the test period.

**Twin-Delayed DDPG**

Here, the TD3 equation is elaborated. Similar to the DDPG, TD3 concurrently learns two Q-functions, $Q_{\phi_1}$ and $Q_{\phi_2}$, to minimize the mean square, Bellman error (MSBE) is defined as a loss function. A clipped noise should be added to each dimension of the actions to form the Q-learning target relying on the target policy $\mu_{\theta_{targ}}$. Then the target action is bounded to the valid action range (Equation S21).

$$a'(s') = \text{clip}\left(\mu_{\theta_{targ}}(s') + \text{clip}(\grave{o}, -c, c), a_{Low}, a_{High}\right), \quad \grave{o} \sim N(0, \sigma) \quad \text{S21}$$

This process, also called target policy smoothing, serves as a regularizer of the algorithm and resolves a particular failure mode happening in the DDPG. In other words, TD3 smooths out the Q-function over similar actions, which is what target policy smoothing is intended to do. Both Q-functions employ a single target, which is calculated by utilizing the Q-function that produces a smaller target value (Equation S22). Then, both Q-functions are learned by regressing to the selected target, which helps fend off overestimation in the Q-function (Equations S23 and S24). Thus, overestimation in the Q-function can be avoided by using the smaller Q-value for the target and regressing towards it.

$$y(r, s', d) = r + \gamma(1-d)\min_{i=1,2} Q_{\phi_{i,targ}}(s', a'(s')) \quad \text{S22}$$

$$L(\phi_1, D) = \underset{(s,a,r,s',d)\sim D}{E}\left[\left(Q_{\phi_1}(s,a) - y(r,s',d)\right)^2\right] \quad \text{S23}$$

$$L(\phi_2, D) = \underset{(s,a,r,s',d)\sim D}{E}\left[\left(Q_{\phi_2}(s,a) - y(r,s',d)\right)^2\right] \quad \text{S24}$$

Finally, similar to the DDPG, the policy would be learned by maximizing $Q_{\phi_1}$ in Equation S25. However, in contrast to the Q-functions, the policy updates are much less frequent in the TD3. This helps mitigate the volatility that frequently occurs in the DDPG as a result of how a policy update alters the target.

$$\max_{\theta} \underset{s\sim D}{E}\left[Q_{\phi_1}(s, \mu_{\theta}(s))\right] \quad \text{S25}$$

Similar to the DDPG, an uncorrelated mean-zero Gaussian noise should be added to the actions during the training to improve TD3 policy exploration.

**Soft Actor Critic**

To begin explaining SAC, the entropy-regularized RL environment is first discussed. The formula for value functions in entropy-regularized RL is slightly different. Entropy is a measure of how unpredictable a random variable is. If a coin is weighted such that it usually always comes up heads, it has a low entropy; if it is uniformly weighted, it has a 50% chance of either outcome. Considering $x$, a random variable with a probability mass function or density function $P$. The entropy $H$ of $x$ is calculated based on its distribution $P$ using Equation S26.

$$H(P) = \underset{x\sim P}{E}\left[-\log P(x)\right] \quad \text{S26}$$

In the entropy-regularized RL, an agent receives a bonus reward at each time step proportionate to the entropy of the policy at that time step. As a result, the RL problem becomes:

$$\pi^* = \arg\max_{\pi} \underset{\tau\sim\pi}{E}\left[\sum_{t=0}^{\infty}\gamma^t\left(R(s_t, a_t, s_{t+1}) + \alpha H\left(\pi(\cdot | s_t)\right)\right)\right] \quad \text{S27}$$

Where $\alpha > 0$ denotes the trade-off coefficient. In this new setting, the value functions are defined slightly different, and the entropy rewards from each time step are included in $V^\pi$ as follows:

$$V^\pi(s) = \mathop{\mathrm{E}}_{\tau \sim \pi}[\sum_{t=0}^\infty \gamma^t \big(R(s_t, a_t, s_{t+1}) + \alpha H\big(\pi(\cdot | s_t)\big)\big) \Big| s_0 = s] \qquad \text{S28}$$

In addition, $Q^\pi$ is updated to consider entropy rewards from each time step except the first:

$$Q^\pi(s, a) = \mathop{\mathrm{E}}_{\tau \sim \pi}[\sum_{t=0}^\infty \gamma^t R(s_t, a_t, s_{t+1}) + \alpha \sum_{t=1}^\infty \gamma^t H\big(\pi(\cdot | s_t)\big) \Big| s_0 = s, a_0 = a] \qquad \text{S29}$$

Considering the above definitions, $V^\pi$ and $Q^\pi$ are associated with Equation S30 and the Bellman equation for $Q^\pi$ is presented by Equation S31:

$$V^\pi(s) = \mathop{\mathrm{E}}_{a \sim \pi}[Q^\pi(s, a) + \alpha H\big(\pi(\cdot | s)\big)] \qquad \text{S30}$$

$$Q^\pi(s, a) = \mathop{\mathrm{E}}_{\substack{s' \sim P \\ a' \sim \pi}}[R(s, a, s') + \gamma\big(Q^\pi(s', a') + \alpha H\big(\pi(\cdot | s')\big)\big)]$$
$$= \mathop{\mathrm{E}}_{s' \sim P}[R(s, a, s') + \gamma V^\pi(s')] \qquad \text{S31}$$

SAC simultaneously learns a policy $\pi_\theta$, as well as two Q-functions, $Q_{\phi_1}$ and $Q_{\phi_2}$. SAC is now available in two versions: one with a fixed entropy regularization coefficient, $\alpha$ (SAC18), and another one with an entropy constraint enforced by varying $\alpha$ over the course of training (SAC19). In this study, both versions were employed to examine their performance on the operating reservoir problem. In both versions, Q-functions are learned in a similar manner to the TD3 with a few exceptions including (i) a term from entropy regularization is also included in the target, (ii) instead of a target policy, the target uses next-state actions from the current policy, (iii) as SAC adopts a stochastic policy, there is no explicit smoothing of target policy and the noise generated by that stochasticity is enough to provide the same result (TD3 trains a deterministic policy, therefore it smooths the next-state actions by injecting random noise).

In order to fully understand where the Q-loss comes from, let's first see how entropy regularization contributes to Q-loss. Considering the recursive Bellman equation for the entropy-regularized $Q^\pi$:

$$Q^\pi(s, a) = \mathop{\mathrm{E}}_{\substack{s' \sim P \\ a' \sim \pi}}[R(s, a, s') + \gamma\big(Q^\pi(s', a') + \alpha H\big(\pi(\cdot | s')\big)\big)]$$
$$= \mathop{\mathrm{E}}_{\substack{s' \sim P \\ a' \sim \pi}}[R(s, a, s') + \gamma\big(Q^\pi(s', a') - \alpha \log \pi(a' | s')\big)] \qquad \text{S32}$$

The right-hand side is an expectation over the next states, which comes from the replay buffer and the next actions that comes from the current policy. Thus, it can be approximated by sampling as it is an expectation:

$$Q^\pi(s, a) \approx r + \gamma\big(Q^\pi(s', \tilde{a}') - \alpha \log \pi(\tilde{a}' | s')\big), \quad \tilde{a}' \sim \pi(\cdot | s') \qquad \text{S33}$$

For each Q-function, SAC calculates the MSBE loss by approximating the target with the samples. Similar to TD3, SAC also employs a clipped double-Q trick and takes the smaller Q-value between the two Q approximators. To summarize, considering the target (Equation S34), the loss function for the Q-functions can be formulated as follows (Equation S35).

$$y(r,s',d) = r + \gamma(1-d)\left(\min_{j=1,2} Q_{\phi_{\text{targ},j}}(s',\tilde{a}') - \alpha \log \pi_\theta(\tilde{a}'|s')\right), \quad \tilde{a}' \sim \pi_\theta(\cdot|s') \tag{S34}$$

$$L(\phi_i, D) = \underset{(s,a,r,s',d)\sim D}{E}\left[\left(Q_{\phi_i}(s,a) - y(r,s',d)\right)^2\right] \tag{S35}$$

The Policy in each state should aim to maximize expected future rewards and entropy. In other words, it should maximize the $V^\pi(s)$ presented below:

$$\begin{aligned}V^\pi(s) &= \underset{a\sim\pi}{E}[Q^\pi(s,a) + \alpha H(\pi(\cdot|s))] \\ &= \underset{a\sim\pi}{E}[Q^\pi(s,a) - \alpha \log \pi(a|s)]\end{aligned} \tag{S36}$$

The reparameterization approach is applied to optimize the policy, which computes a deterministic function of the state, policy parameters, and independent noise to draw a sample from $\pi_\theta(\cdot|s)$. To do so, a squashed Gaussian policy (e.g., *tanh*) is considered by the authors to draw samples:

$$\begin{aligned}V^{\pi_\theta}(s) &= \underset{a\sim\pi_\theta}{E}[Q^{\pi_\theta}(s,a) + \alpha H(\pi_\theta(\cdot|s))] \\ &= \underset{a\sim\pi_\theta}{E}[Q^{\pi_\theta}(s,a) - \alpha \log \pi_\theta(a|s)]\end{aligned} \tag{S37}$$

In contrast to vanilla policy gradient (VPG), trust region policy optimization (TRPO), and PPO, the *tanh* function employed in the SAC policy squashes all actions to a specific range. The SAC reparameterization trick makes it possible to rewrite the expectation over actions (the distribution relies on the policy parameters which is a pain point) into an expectation over noise (the distribution no longer depends on parameters).

$$\underset{a\sim\pi_\theta}{E}[Q^{\pi_\theta}(s,a) - \alpha \log \pi_\theta(a|s)] = \underset{\xi\sim N}{E}[Q^{\pi_\theta}(s,\tilde{a}_\theta(s,\xi)) - \alpha \log \pi_\theta(\tilde{a}_\theta(s,\xi)|s)] \tag{S39}$$

The next step is to substitute $Q^{\pi_\theta}$ with one of the function approximators (minimum of the two Q approximators) in order to obtain the policy loss. Therefore, the policy would be optimized based on Equation S40 which is similar to the DDPG and TD3 policy optimizations (except for the min-double-Q trick, the stochasticity, and the entropy term).

$$\max_\theta \underset{\substack{s\sim D \\ \xi\sim N}}{E}[\min_{j=1,2} Q_{\phi_j}(s,\tilde{a}_\theta(s,\xi)) - \alpha \log \pi_\theta(\tilde{a}_\theta(s,\xi)|s)] \tag{S40}$$

As mentioned, SAC is an on-policy algorithm that trains a stochastic policy leveraging entropy regularization. The entropy regularization coefficient, $\alpha$, controls the trade-off between exploration and exploitation (a higher $\alpha$ value indicates more exploration and vice versa). To determine how well the policy exploits the learned information during the testing period, the stochasticity should be removed, and the actions mean should be used instead of sampling from the distribution.